\pdfoutput=1
\documentclass[11pt]{article}

\usepackage{acl}
\usepackage{times}
\usepackage{latexsym}

\usepackage[T1]{fontenc}
\usepackage[utf8]{inputenc}

\usepackage{microtype}
\usepackage{epsfig}
\usepackage{graphicx}
\usepackage{amsmath}
\usepackage{amssymb}
\usepackage{amsfonts}
\usepackage{bbm}
\usepackage[normalem]{ulem} 
\usepackage{booktabs}
\usepackage{multirow}
\usepackage{algorithm}
\usepackage[noend]{algorithmic}
\usepackage{makecell}
\usepackage{enumitem}
\usepackage{epigraph}
\usepackage{csquotes}
\usepackage{mathtools}
\usepackage{textgreek}
\usepackage{color, colortbl}
\usepackage{sepfootnotes}

\definecolor{LightCyan}{rgb}{0.88,1,1}
\definecolor{LightRed}{rgb}{1.0, 0.91, 0.91}
\definecolor{LightGray}{rgb}{0.88,0.88,0.88}
\definecolor{VeryLightGray}{rgb}{0.93,0.93,0.93}
\makeatletter
\newcommand{\printfnsymbol}[1]{%
  \textsuperscript{\@fnsymbol{#1}}%
}

\DeclareMathOperator*{\E}{\mathbb{E}}
\DeclareMathOperator{\R}{\mathbb{R}}
\DeclareMathOperator{\x}{\mathbf{x}}
\DeclareMathOperator{\y}{\mathbf{y}}

\DeclareMathOperator*{\argmax}{argmax}

\DeclarePairedDelimiter\floor{\lfloor}{\rfloor}
\newcommand{\textul}{\uline}

\newcommand*\samethanks[1][\value{footnote}]{\footnotemark[#1]}

  {\par\kern\topsep}% \end{mycenter}

\begin{document}
%%%%%%%%% TITLE
\title{Semantically Distributed Robust Optimization \\for Vision-and-Language Inference}
\newcounter{thanks}
\author{
	Tejas Gokhale\thanks{~~Equal Contribution},
	~~Abhishek Chaudhary\samethanks,\\
	{\bf
	Pratyay Banerjee,~~Chitta Baral, \and 
	Yezhou Yang
	} \\
	Arizona State University \\
    {\tt\small \{tgokhale, achaud39, pbanerj6, chitta, yz.yang\}@asu.edu}
}

\maketitle

%%%%%%%%% ABSTRACT
\setcounter{footnote}{1}
\begin{abstract}
Analysis of vision-and-language models has revealed their brittleness under linguistic phenomena such as paraphrasing, negation, textual entailment, and word substitutions with synonyms or antonyms.
While data augmentation techniques have been designed to mitigate against these failure modes, methods that can integrate this knowledge into the training pipeline remain under-explored.
In this paper, we present \textbf{SDRO}\footnote{\href{https://github.com/ASU-APG/VLI_SDRO}{\footnotesize\url{https://github.com/ASU-APG/VLI_SDRO}}}, 
a model-agnostic method that utilizes a set linguistic transformations in a distributed robust optimization setting, along with
an ensembling technique to leverage these transformations during inference.
Experiments on benchmark datasets with images (NLVR$^2$) and video (VIOLIN) demonstrate performance improvements as well as robustness to adversarial attacks.
Experiments on binary VQA explore the generalizability of this method to other V\&L tasks.

\end{abstract}
%-------------------------------------------------------------------------
%%%%%%%%% INTRODUCTION
\section{Introduction}
\label{sec:01_intro}
\begin{figure}[t]
    \centering
    \includegraphics[width=\linewidth]{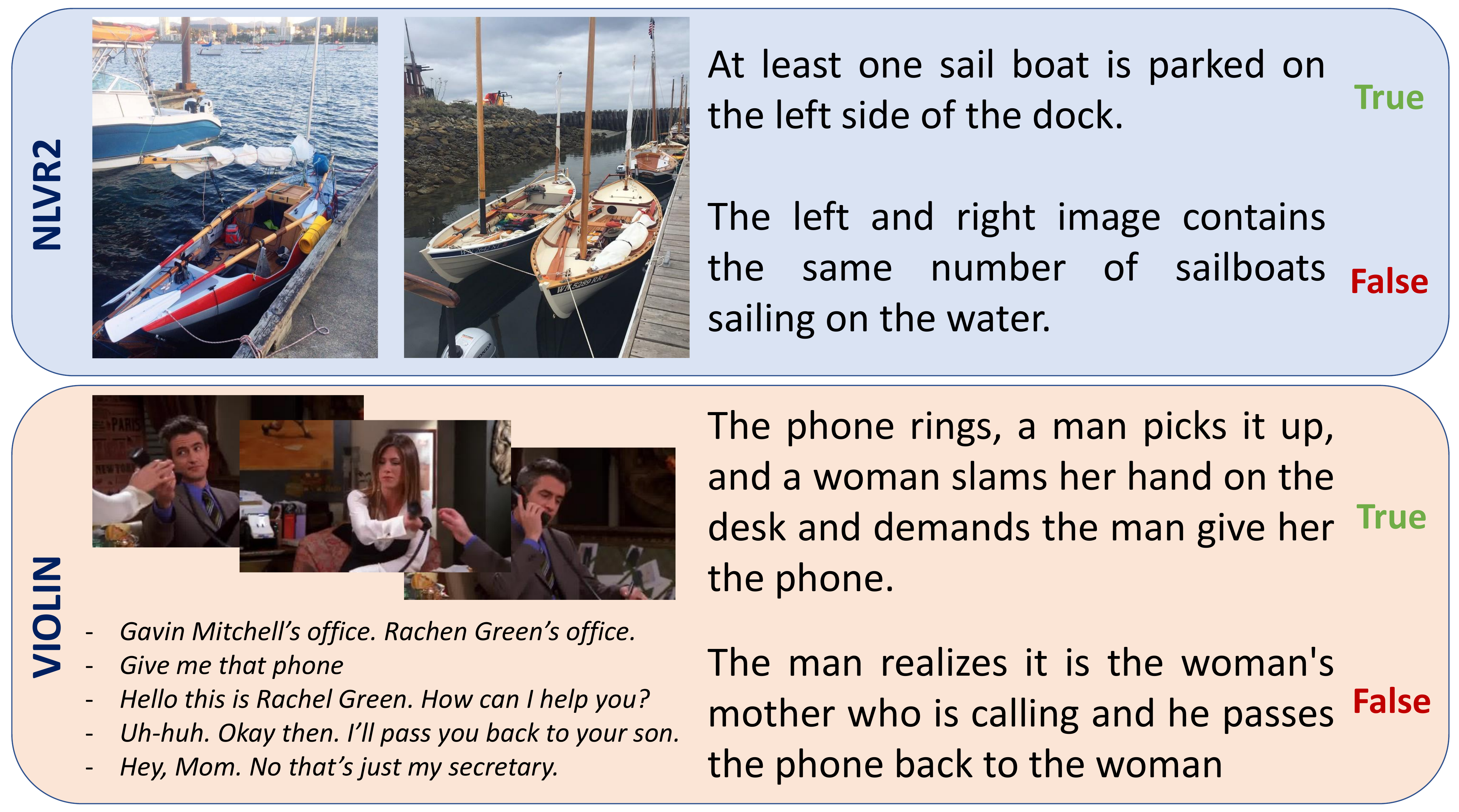}
    \caption{
        VLI models predict whether a sentence is \texttt{True} or \texttt{False}, given the visual input. 
        \textit{(Top)} sample from NLVR$^2$ with two images as input;
        \textit{(bottom)} sample from VIOLIN with video and subtitles as input.
        }
    \label{fig:example_vli}
\end{figure}

\hfill{\centering\textit{``Does the text match the image?''}}\\
-- this simple question represents the Vision-and-Language Inference (VLI) task,
as shown in Figure~\ref{fig:example_vli}.
Image-text matching forms the backbone for V\&L pre-training~\cite{sun2019videobert,tan2019lxmert,lu2019vilbert} and has resulted in improvements in downstream tasks such as visual question answering, image retrieval, referring expressions, and visual commonsense reasoning.
While natural language inference (without visual inputs) has been extensively studied~\cite{bowman2015large,williams2018broad,khot2018scitail,demszky2018transforming}, VLI demands the additional capability of being grounded in the scene while understanding semantics.
Although pre-trained language models (PLMs)~\cite{vaswani2017attention,devlin2019bert,raffel2020exploring} have been useful for encoding text into vector embeddings, recent findings point to undesirably high cosine similarity of two random words~\cite{ethayarajh2019contextual}, the struggle with negation~\cite{kassner2020negated,ettinger2020bert}, and semantically equivalent adversarial examples~\cite{ribeiro2018semantically}.
These findings call for robust training protocols to avoid propagation of these findings into VLI models.

Adversarial training (AT) and distributed robust optimization (DRO)~\cite{madry2018towards,hu2018does,sinha2017certifying} have emerged as effective solutions to related problems in robust image classification, such as adversarial defense and domain generalization~\cite{volpi2018generalizing}.
DRO assumes a perturbation set (typically an $\ell_p$ norm ball) around the training distribution, and minimizes the worst-case performance over this perturbation set.
AT and DRO are popular for computer vision tasks, since the small perturbations of pixel intensities do not change the categorical meaning of the image.

\begin{figure}
    \centering
    \includegraphics[width=\linewidth]{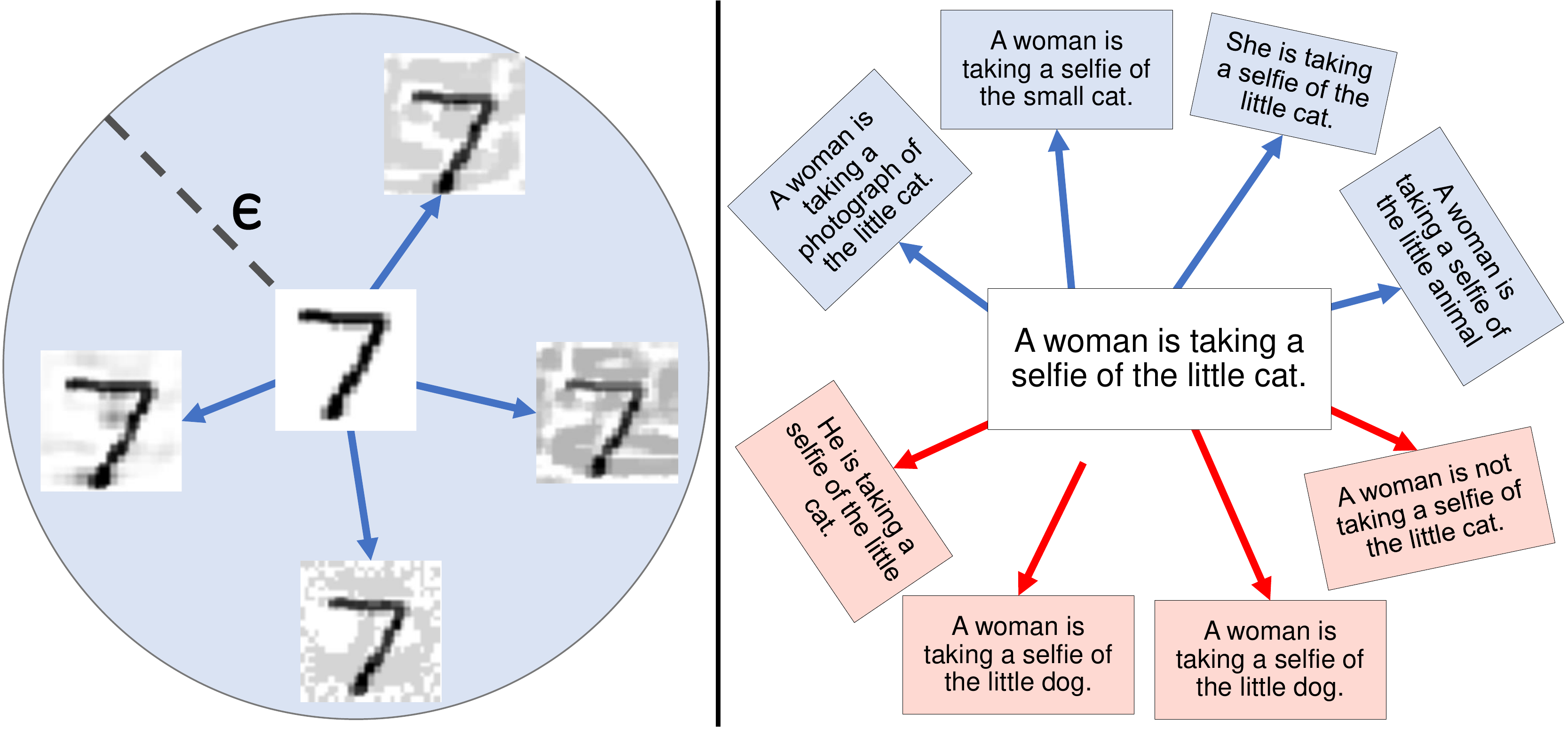}
    \caption{
    Comparison between \textit{(left)}
    \straightepsilon-bounded image perturbations and \textit{(right)} linguistics-based \textit{semantics-preserving} (blue) as well as  \textit{semantics-inverting} (red) transformations for sentences.
    }
    \label{fig:vector_vs_linguistic}
\end{figure}

However, in the case of text inputs, even small perturbations of their vector embeddings may result in absurd sentences or vectors that do not map to any word-token in vocabulary. 
The topology of the PLM embedding space is not well understood, especially with regard to what kind (and magnitude) of perturbations result in specific changes in semantics, such as similar meanings (speak $\rightarrow$ talk) or opposite meanings (Heaven $\rightarrow$ Hell) without resulting in random or absurd words.
Vector-based additive perturbations of text inputs thus restrict interpretability.
However, in the domain of natural language, knowledge of logic, grammar, and semantics can be leveraged to transform sentences as shown in Table~\ref{tab:sisp_examples}.
Such linguistically-informed perturbations provide us control over the semantics of the resulting sentence and label, as shown in Figure~\ref{fig:vector_vs_linguistic}.

We present a technique that modifies robust optimization by incorporating linguistically-informed transformations.
Our approach: \textbf{S}emantically \textbf{D}istributed \textbf{R}obust \textbf{O}ptimization (SDRO) 
utilizes a pre-defined set of linguistic transformations (such as negation, word substitution, and paraphrasing) as the perturbation set instead of optimizing over the vector-space.
We dub this set of transformations ``SISP'' i.e., semantics-inverting (SI) and semantics-preserving (SP) transformations.
SDRO is \textit{model-agnostic} since it can be applied to text inputs of any existing VLI model and \textit{dataset agnostic} since it uses automated transformations without explicit knowledge of the text domain.

We apply SDRO to two VLI benchmark datasets: image-based NLVR$^2$~\citep{suhr2019corpus} as well as video-based VIOLIN~\citep{liu2020violin}.
To demonstrate the generalizability of SDRO to other V\&L tasks, we also report results on the ``yes/no'' subset of VQA-v2~\cite{goyal2017making}.
Our experiments show model-agnostic improvements in accuracy for all three benchmarks.
While models trained with naive data augmentation using SISP suffer from a trade-off between robustness and accuracy, models that utilize SDRO improve along both metrics.
SDRO also allows us to learn in low-resource settings, serving as a smart data augmentation tool -- SDRO models trained only with $80\%$ of the original dataset outperform existing state-of-the-art which utilizes the entire dataset.

Since SISP transforms do not require the ground truth label to either produce an SP or SI transformed sentence, we can also apply them to any new input sentences that are observed at test-time.
Given a test input sentence, we generate its SISP versions, and obtain the prediction from our model for each SISP version.
These predictions are ensembled using weighted averaging, giving equal weight to the prediction for the original sentence and the average predictions for all transformed sentences.
We find that this ensembling of predictions of the SDRO model at test-time pushes the state-of-the-art further, thereby demonstrating the usefulness of semantic sentence transformations, both during training and testing.

%-------------------------------------------------------------------------
%%%%%%%%% SDRO
\section{Method} 
\begin{table*}[t]
    \centering
    \small
    \resizebox{\linewidth}{!}{
    \begin{tabular}{@{}clp{0.42\linewidth}p{0.43\linewidth}@{}}
        \toprule
         & \textbf{Category}    & \textbf{Original} & \textbf{Transformed} \\
        \toprule 
        \multirow{10}{*}{\rotatebox{90}{SI}}
         & Noun-Antonym
            & The two women are driving on the street with the convertible top down.
            & The two \textit{\textbf{men}} are driving on the street with the convertible top down.\\
         & Verb-Antonym
            & There are children standing by the door.
            & There are children \textbf{\textbf{sitting}} by the door.\\
         & Comparative-Antonym
            & There are more monitors in the image on the right than on the left.
            & There are \textit{\textbf{few}} monitors in the image on the right than on the left. \\
         & Number-Substitution
            & There are three bowls of dough with only one spatula.
            & There are \textit{\textbf{eleven}} bowls of dough with only one spatula. \\
         & Pronoun-Substitution 
            & In one of the images, a woman is taking a selfie. 
            & In one of the images, \textit{\textbf{he}} is taking a selfie.\\
         & Subject-Object Swap
            & The two women are driving on the street with the convertible top down. 
            & The two \textit{\textbf{top}} are driving on the street with the convertible \textit{\textbf{women}} down.\\
         & Negation
            & The closet doors on the right are mirrored.
            & The closet doors on the right are \textit{\textbf{not}} mirrored\\
        \midrule
        \multirow{8}{*}{\rotatebox{90}{\footnotesize SP}}
         & Noun-Synonym         
            & The right image shows three bottles of beer lined up.
            & The right \textit{\textbf{picture}} shows three bottles of beer lined up.\\
         & Verb-Synonym         
            & Someone is using a kitchen utensil
            & Someone is \textit{\textbf{utilizing}} a kitchen utensil. \\
         & Comparative-Synonym  
            & The bottle on the right is larger than the bottle on the left.
            & The bottle on the right is \textit{\textbf{bigger}} than the bottle on the left.\\
         & Number-Substitution  
            & The two white swans are swimming in the canal gracefully. 
            & The \textit{\textbf{less than seven}} white swans are swimming in the canal gracefully.\\ 
         & Pronoun-Substitution 
            & In one of the images, a woman is taking a selfie. 
            & In one of the images, \textit{\textbf{she}} is taking a selfie.\\
         & Paraphrasing         
            & A man in a green shirt came on the porch and knocked on the door.
            & A man in a green shirt came \textit{\textbf{up to}} the porch and knocked on the door. \\
        \bottomrule 
    \end{tabular}
    }
    \caption{Examples illustrating the effect of each SISP transformation on input sentences.}
    \label{tab:sisp_examples}
\end{table*}

\subsection{Preliminaries}
Consider a training distribution $P_{tr}$ consisting of inputs $\x$ and labels $\y$.
For VLI, input $\x$ is multi-modal (visuals and text), with labels $\y \in \{\texttt{True}, \texttt{False}\}$.
Under the empirical risk minimization (ERM), the following risk is minimized:
\begin{equation}
    \mathcal{R}_{ERM} = \E_{(\x, \y) \sim P_{tr}}  ~\ell(f(\x; \theta), \y),
    \label{eq:erm}
\end{equation}
where $\ell$ is a suitable loss function such as cross-entropy loss for classification tasks.
ERM provides generalization guarantees~\cite{vapnik1991principles} for i.i.d.\ test samples, but not for out-of-distribution or adversarial examples.

\paragraph{Distributed Robust Optimization (DRO)}~\cite{pmlr-v80-hu18a,sagawa2020distributionally} searches for loss-maximizing perturbations of the input within an $\epsilon$-divergence ball around $P_{tr}$ and minimize the risk over such perturbed distributions.
\begin{equation}
    \mathcal{R}_{DRO}=\underset{P:D(P, P_{tr})<\epsilon}{sup}\E_{(\x, \y) \sim P} \ell(f(\x; \theta), \y).
    \label{eq:std_adv}
\end{equation}
The solution to Equation~\ref{eq:std_adv} guarantees robustness inside such $\epsilon$-bounded distributions $P$.
The inner maximization is typically solved using gradient-based methods~\citep{madry2018towards} over additive perturbations $\delta$ such that $\x+\delta$ fools the classifier.

\subsection{SDRO}
\label{sec:sdro}
For sentence inputs, additive perturbations are intangible and may result in ambiguity.
An alternative approach is to consider \textit{groups} $\mathcal{G}$ representing certain sub-populations or semantic categories within the data distribution.
For text inputs in VLI, we consider the use of semantic sentence transformations as the perturbation mechanism -- thus each transformation creates a sub-population or group of sentences.
Examples of these transformations and their resulting effect on sentences is shown in Table~\ref{tab:sisp_examples}.
% we consider perturbation sets that can be created using semantic sentence transformations such as those shown in Table~\ref{tab:sisp_examples}, as the \textit{groups} (or equivalently, \textit{transformations}).
These transformations $g(x, y) = (\x_g, \y_g)$ are of two types:
semantics-preserving (SP) if $\y_g{=}\y$, or semantics-inverting (SI) if $\y_g \neq \y$.

In this paper we propose the use of SI and SP transformations to create groups within the training data which can be leveraged by a robust optimization techniques to minimize worst group error.
While previous work on adversarial training uses vector perturbations of sentence embeddings, our sentence-level transformations are interpretable 
% (since the effect of the transformation can be mapped back to natural language, 
as shown in Table~\ref{tab:sisp_examples}).
The ability of generating adversarial samples with inverted meanings is a key distinction between adversarial training (AT) and SDRO.
While AT is restricted to SP perturbations inside an $\epsilon$ norm-ball, SDRO can impart larger linguistic perturbations (both SI and SP) beyond the norm-ball, by minimizing the worst-case expected risk over these groups:
\begin{equation}
    \mathcal{R}_{SDRO} = \underset{g\in\mathcal{G}}{sup}\E_{(\x, \y) \sim g} ~\ell(f(\x; \theta), \y).
    \label{eq:sdro}
\end{equation}

\paragraph{Implementation.}
As a first step of SDRO, we randomly sample a subset $\mathcal{C}$ of the training dataset $\mathcal{D}$ s.t. $|\mathcal{C}|/|\mathcal{D}|=T$. 
We find adversarial samples after every epoch and create an augmented dataset $\mathcal{D}_{aug}$ which contains $(1-T)|\mathcal{D}|$ original samples and $T|\mathcal{D}|$ adversarial samples, thus retaining the size of the training dataset.
We define $\ell_g$ as the classification loss for a transformed sample $(\x_g, \y_g)$:
\begin{equation}
    \ell_g(\x, \y) \triangleq \ell(f(\x_g), \y_g),~\forall g{\in}\mathcal{G}.
    \label{eq:ell_G}
\end{equation}

\subsection{Variants of SDRO}
We design two variants of SDRO: Sample-Wise (SW) and Group-Wise (GW) SDRO.
\paragraph{Sample-Wise SDRO} is a greedy version of SDRO, in which, for every input $\x$, a transformation that maximally fools the classifier: $g^*{=}\argmax_{g\in\mathcal{G}} \ell_g(\x, \y)$, is added to the set of adversarial examples $\mathcal{D}_{adv}$. 
The model is then fine-tuned on the augmented dataset.
\begin{align}
    \mathcal{D}_{adv} &= \{ g^*(\x, \y) \colon (\x, \y){\in}~\mathcal{C}\}, \\
    \mathcal{D}_{aug} &= \mathcal{D}_{1:(1-T)|\mathcal{D}|} \cup \mathcal{D}_{adv}
    \label{eq:d_aug}
\end{align}
However, this greedy approach is susceptible to the model's biases towards certain transformations.  
For instance, if negation and verb-antonym are universally hard for most sentences, i.e., result in the maximum classifier loss amongst all  transformations $g$, then $\mathcal{D}_{adv}$ will be dominated by these groups, resulting in an unbalanced training set.

\paragraph{Group-Wise SDRO}
is devised to mitigate against the model becoming biased towards the ``hardest'' transformations.
Using Equation~\ref{eq:ell_G}, we calculate the transformation losses for each transformation of each sample in a training batch, yielding a set of classifier losses per ``group'' $g$:
\begin{equation}
    L_g \colon \mathcal{C} \rightarrow \R;
    ~L_g = \{ \ell_g(\x, \y) \colon (\x, \y) \in \mathcal{C}\}.
\end{equation}
We obtain the top-k losses per group $g$ as:
\begin{equation}
    L_G^k = \argmax_{\Lambda \subset L_G, |\Lambda| = k} \sum_{\lambda\in\Lambda} \lambda,
    ~~\text{where~}k = \floor*{\frac{|\mathcal{C}|}{|\mathcal{G}|}}.
    \label{eq:pergroup_losses}
\end{equation}
Then 
$\mathcal{D}_{adv}$ is compiled as the union of per-group adversaries using Equation~\ref{eq:pergroup_losses}, and augmented to the training dataset using Equation~\ref{eq:d_aug}.
    
\paragraph{Test-Time Ensembling of Predictions.}
Semantic transformations $g$ allow us to obtain multiple ``views'' $\x_g = g(\x)$ of the input, and the corresponding predictions $\hat{\y}_g = f(\x_g)$.
We ensemble these predictions and the original prediction $\hat{\y}=f(\x)$ with a simple weighted-average.
Note that $\mathcal{G}$ contains both SP and SI transformations, $\mathcal{G}_{SP}$ and $\mathcal{G}_{SI}$. 
Since the expected label for $\mathcal{G}_{SI}$ is flipped, during ensembling we use the flipped probabilities $1{-}f(\x_g)$.
The ensembled prediction is:
\begin{equation}
\hat{\y}_{e} = \alpha f(\x) +
\frac{1{-}\alpha}{2}\sum_{\mathclap{g\in\mathcal{G}_{SP}}}{\frac{f(\x_g)}{|\mathcal{G}_{SP}|}} + \frac{1{-}\alpha}{2}\sum_{\mathclap{g\in\mathcal{G}_{SI}}}\frac{1{-}f(\x_g)}{|\mathcal{G}_{SI}|}.
\small
% \hat{\y}_{e} = \alpha f(\x) + \frac{1{-}\alpha}{2}\E_{g\in\mathcal{G}_{SP}}{f(\x_g)} + \E_{g\in\mathcal{G}_{SI}}{1{-}f(\x_g)}.
\label{eq:ensembling}
\end{equation}
Note that our method is a test-time ensemble and does not require training multiple models.
This method is, in principle, similar to the ensembling strategy in image classification used by \citet{chai2021ensembling} who train a generative model $g$ to output different views of an image, and tune $\alpha$ over a validation set.
In our work, $g$ are semantic sentence transformations, and the value of $\alpha$ does not need to be tuned over a validation set -- we find that the simple intuitive choice of $\alpha{=}0.5$ (equal weight to the original sample and the SISP versions) improves performance.
We find that:
\begin{itemize}[nosep,noitemsep]
\item training models with SDRO using SISP transformations improves results on VLI tasks, and
\item ensembling predictions of SDRO at test-time using Equation~\ref{eq:ensembling} further improves results.
\end{itemize}
%-------------------------------------------------------------------------
\section{SISP Sentence Transformations}
This section describes the generation of semantics-preserving (SP) and semantics-inverting (SI) statements.
\textbf{SISP} transforms are implemented using Spacy~\cite{spacy}.
Dataset statistics and additional visualizations are in the Appendix.

\paragraph{Noun Synonym/Antonym:}
We extract nouns (subjects and objects) with dependency parsing, and find two nearest (synonyms) or farthest (antonyms) neighbors in the GloVe space~\cite{pennington2014glove} using a threshold of $0.55$.

\paragraph{Verb Synonym/Antonym:}
We extract verbs using POS tagging and obtain their synonyms or antonyms.
Verbs are lemmatized and inflected to the correct form using Lemminflect~\cite{lemminflect}.

\paragraph{Comparative Synonym/Antonym:}
Adjectival complements and modifiers are replaced with synonyms (\textit{large $\rightarrow$ big}) or antonyms (\textit{large $\rightarrow$ small}). 

\paragraph{Number Substitution:}
Numerals are replaced by number-words (2 $\rightarrow$ \textit{two}) or vice versa for SP transformations, or by their lower or upper bounds,
(SP: \textit{3 $\rightarrow$ more than two}, SI: \textit{two $\rightarrow$ less than two}).

\paragraph{Pronoun Substitution:}
Human-related nouns (such as \textit{woman, boy, people}) are substituted by pronouns, while pronouns are substituted by generic descriptors (\textit{something, someone, somebody, they}).

\paragraph{Negation:}
We use template-based negation~\cite{gokhale2020vqa} with Subject-Verb Agreement~\cite{wren2000english}.
We add \textit{`did not'} before a past-tense verb, \textit{`do not'}, \textit{`does not'}, or \textit{`not'} before a base-form verb, gerund, or participle, or a \textit{`not'} before an adposition or adjective.

\paragraph{Subject-Object Swap:}
Nominal or clausal subjects and direct or prepositional objects from the sentence are swapped for inverting semantics.

\paragraph{Paraphrasing:}
Input sentences are translated to Russian and then back-translated to English using neural machine translation~\cite{ott2019fairseq}.

\subsection{Data Analysis}
\label{sec:sisp_fidelity_bias}
\paragraph{Quantification of Bias:}
\begin{table}[t]
    \centering
    \footnotesize
    \resizebox{\linewidth}{!}{
    \begin{tabular}{@{}l ccc c ccc@{}}
        \toprule
        \multirow{2}{*}{\textbf{Method}} & \multicolumn{3}{c}{NLVR2} & \hphantom & \multicolumn{3}{c}{VIOLIN} \\ 
         \cmidrule{2-4} \cmidrule{6-8}
         & Clean & SP & SI && Clean & SP & SI \\
         \midrule 
        Data-Aug & 51.07 & 50.92 & 40.74 && 61.12 & 62.78 & 62.15 \\
        SW-SDRO  & 51.14 & 50.97 & 40.75 && 62.78 & 58.13 & 64.78 \\
        GW-SDRO  & 51.07 & 50.92 & 40.73 && 62.15 & 52.79 & 74.98 \\
        \bottomrule
    \end{tabular}
    }
    \caption{Text-only evaluation of biases due to SISP transformations. $50\%$ indicates no bias.}
    \label{tab:eval_textonly_bias}
\end{table}
Since SISP transforms are based on templates, they can potentially introduce spurious linguistic correlations in the dataset.
For example, in NLVR$^2$ and VIOLIN datasets, negations and indefinite pronouns
are infrequent.
To quantify how this could impact models, we mask out the entire image and evaluate models (with VILLA as the backbone for NLVR$^2$ and HERO for VIOLIN).
This acts as a `text-only' evaluation, with accuracies ${\sim}50\%$ implying lesser bias since models do not have access to visual information.
Table~\ref{tab:eval_textonly_bias} shows that SP transforms inflict lesser bias on models than SI transforms.
The effect of bias is dataset-specific; SI makes the prediction of NLVR$^2$ samples harder than random (less than $50\%$ accuracy) but easier for VIOLIN.

\paragraph{Transformation Fidelity:}
We employ human subjects to evaluate the quality of SISP-transformed sentences on (1) correctness of labels, (2) grammar, (3) semantics, and (4) visual grounding. 
We report a unified average `transformation fidelity' (details are in Appendix).
Fidelity is higher for SP samples than SI ($90.50\%$ v/s $79.51\%$), which resonates with the complexities of inversion of meaning~\cite{russell1905denoting} and leaves room for improvement in SI transformation.

%-------------------------------------------------------------------------
%%%%%%%%% EXPERIMENTS
\section{Experiments}
\paragraph{Datasets.}
For all datasets, given images/videos and natural language text as input, the system is expected to predict a binary class label.
NLVR$^2$~\cite{suhr2019corpus} contains ${\sim}86K, 7K, 7K$ samples for training, development, and testing respectively.
Each sample in NLVR$^2$ consists of a pair of images (from search engines) and a sentence (crowd-sourced).
VIOLIN~\cite{liu2020violin} contains video clips from popular TV shows and movies along with subtitles and crowd-sourced statements.
VIOLIN contains $76K$, $9.5K$, $9.5K$ samples for training, validation and testing.
VQA Yes/No consists of image-question-answer triplets from VQA-v2 dataset~\cite{goyal2017making}.
While VQA-v2 consists of multiple question and answer types, we focus on the subset of questions with binary \textit{yes/no} answers (${\sim}38\%$ of VQA-v2).
    
\paragraph{Evaluation Metrics.}
We use two evaluation metrics:
(1) \textbf{Clean Accuracy}: accuracy on the i.i.d.\ benchmark test set, and
(2) \textbf{SISP Accuracy}: average performance on SISP transformations of the test set.
Since SISP transformations are automated and can be noisy (Sec~\ref{sec:sisp_fidelity_bias}), evaluation on the SISP test set can be considered a proxy for robustness.

\footnotetext[3]{
Notation: \textbf{bold}: $>$ SOTA;
shaded: $>$ respective backbone model ($BASE$);
\uline{underlined}: best SI/SP accuracies.
}

\subsection{Results}
We compare SDRO with backbone models that use standard training data \textit{(BASE)} and data-augmentation \textit{($+$data-aug)}.
We train SDRO and backbones with the same hyperparameters.
We apply test-time ensembling to the best SDRO model.
\paragraph{NLVR$^2$:}
\begin{table}[t]
    \centering
    % \footnotesize
    \resizebox{\linewidth}{!}{
    \begin{tabular}{@{}lcccc@{}}
        \toprule
        \multirow{2}{*}{\textbf{Model}} & \multirow{2}{*}{\textbf{Clean Acc.}} & \multicolumn{3}{c}{\textbf{SISP Acc.}} \\
        \cmidrule{3-5}
         &  & SP & SI & Avg. \\
        \midrule
        LXMERT$_{BASE}$                 & 74.37 & 69.20 & 37.35 & 53.28 \\
        \ \ \ + VILLA                   & \cellcolor{VeryLightGray}{75.98} & 69.94 & 39.09 & 56.15\\
        \ \ \ + data-aug         & 71.83 & 70.13 & 66.34 & 68.23 \\
        \ \ \ + SW-SDRO          & 71.19 & 67.41 & 66.32 & 66.86 \\
        \ \ \ + GW-SDRO         & \cellcolor{VeryLightGray}{74.55} & 69.06 & 69.34 & 69.20\\
        \ \ \ \quad + Test-Time Ensembling      & \cellcolor{VeryLightGray}{74.75} & --''-- & --''-- & --''-- \\
        \midrule
        UNITER$_{BASE}$                 & 77.85 & 72.73 & 34.86 & 53.80 \\
        \ \ \ + data-aug         & 76.65 & 70.34 & 81.04 & 75.69 \\
        \ \ \ + SW-SDRO        & \cellcolor{VeryLightGray}{78.43} & 69.71 & 67.50 & 68.61 \\
        \ \ \ + GW-SDRO         & 77.55 & 67.93 & 81.66 & 74.79\\
        \ \ \ \quad + Test-Time Ensembling      & \cellcolor{VeryLightGray}{80.00} & --''-- & --''-- & --''-- \\
        \midrule 
        VILLA$_{BASE}$               & 78.39 & \textul{73.15} & 34.15 & 53.65 \\
        \ \ \ + data-aug         & 78.34 & 72.11 & 84.44 & \textul{77.77} \\
        \ \ \ + SW-SDRO         & \cellcolor{VeryLightGray}{\textbf{79.23}} & 69.23 & 67.35 & 68.29 \\
        \ \ \ + GW-SDRO          & \cellcolor{VeryLightGray}{\textbf{79.41}} & 68.67 & \textul{84.54} & 76.60 \\
        \ \ \ \quad + Test-Time Ensembling      & \cellcolor{VeryLightGray}{\textbf{82.22}} & --''-- & --''-- & --''-- \\
        \bottomrule
    \end{tabular}
    }
    \caption[Results on the NLVR2 public test set.]{
        Results on the NLVR$^2$ public test set.
        \footnotemark
        }
    \label{tab:results_nlvr2}
\end{table}
We use Transformer-based models LXMERT~\cite{tan2019lxmert}, UNITER~\cite{chen2020uniter}, and VILLA~\cite{gan2020large} as backbones for SDRO. 
VILLA (the current state-of-the-art for NLVR$^2$) uses standard adversarial training.
The percentage of SISP-transformed samples is fixed at $T{=}20\%$.
Table~\ref{tab:results_nlvr2} shows results on the NLVR$^2$ test set, with consistent model-agnostic improvements in clean accuracy over each baseline model and improved robustness on average.
Both variants of SDRO improve over VILLA$_{BASE}$ by $0.84\%$ and $1.02\%$, respectively.
Test-time ensembling using Equation~\ref{eq:ensembling} leads to further gains, resulting in a new state-of-the-art accuracy of $82.22\%$, an improvement of $3.83\%$ over VILLA$_{BASE}$.
GW-SDRO results in the highest SI accuracy when used with each backbone model.

\paragraph{VIOLIN:}
\begin{table}[t]
    \centering
    \resizebox{\linewidth}{!}{
    \begin{tabular}{@{}lcccc@{}}
        \toprule
        \multirow{2}{*}{\textbf{Model}} & \multirow{2}{*}{\textbf{Clean Acc.}} & \multicolumn{3}{c}{\textbf{SISP Acc.}} \\
        \cmidrule{3-5}
         &  & SP & SI & Avg. \\
        \midrule
        VIOLIN$_{BASE}$              & 68.07 & 57.17 & 57.20 & 57.18 \\           
        \ \ \ + data-aug            & 61.58 & \textul{67.64} & 67.70 & 67.67 \\
        \ \ \ + SW-SDRO             & 62.81 & 62.84 & 62.68 & 62.76 \\
        \ \ \ + GW-SDRO             & 63.71 & 64.58 & 63.16 & 63.87 \\
        \ \ \ \quad + Test-Time Ensembling     & 66.56 & --''-- & --''-- & --''-- \\
        \midrule
        HERO$_{BASE}$               & 68.55 & 65.59 & 32.00 & 48.80 \\
        \ \ \ + data-aug            & 65.21 & 59.20 & 81.81 & \textul{70.51} \\
        \ \ \ + SW-SDRO    & \cellcolor{VeryLightGray}{\textbf{68.83}} & 58.97 & 77.83 & 68.41 \\
        \ \ \ + GW-SDRO             & 68.19 & 56.20 & \textul{82.92} & 69.57 \\
        \ \ \ \quad + Test-Time Ensembling     & \cellcolor{VeryLightGray}{\textbf{69.90}} & --''-- & --''-- & --''-- \\
        \bottomrule
    \end{tabular}
    }
    \caption[Results on VIOLIN test set.]{Results on VIOLIN test set.\footnotemark[\value{footnote}]}
    \label{tab:results_violin}
\end{table}
We consider VIOLIN$_{BASE}$~\cite{liu2020violin} and HERO~\cite{li2020hero}, the current state of the art, as baselines.
VIOLIN$_{BASE}$ separately computes visual features using Faster-RCNN~\cite{ren2015faster} and textual features using BERT~\cite{devlin2019bert}, and fuses them to be used as input to a classifier model.
On the other-hand, HERO is a large-scale transformer-based pre-trained model which uses various V\&L pre-training tasks to compute cross-modal features.
We set $T{=}40\%$.
The results can be seen in Table~\ref{tab:results_violin}.
SW-SDRO model with the HERO backbone improves the state-of-the-art to $68.83\%$, and test-time ensembling further improves it to $69.90\%$. 
Interestingly, similar improvements in clean accuracy are not observed for VIOLIN$_{BASE}$, potentially because it does not use cross-modal pre-trained features.

\paragraph{VQA Yes/No:}
\begin{table}[t]
    \centering
    \resizebox{\linewidth}{!}{
    \begin{tabular}{@{}lcccc@{}}
        \toprule
        \multirow{2}{*}{\textbf{Model}} & \multirow{2}{*}{\textbf{Clean Acc.}} & \multicolumn{3}{c}{\textbf{SISP Acc.}} \\
        \cmidrule{3-5}
         &  & SP & SI & Avg. \\
        \midrule
        % % LXMERT (all data)           & 86.65 & 75.41 & 34.09 & 54.75\\
        % LXMERT\footnote{LXMERT and UNITER follow a different train-val split}                      & 82.71 & 75.53 & 33.32 & 53.92\\
        % LXMERT\footnotemark[1] + LOL                & 84.75 & 73.77 & 42.97 & 58.37 \\
        % LXMERT-VILLA              & {83.85} & 76.70 & 34.68 & 55.59 \\
        % \midrule
        UNITER$_{BASE}$             & 83.49 & 72.04 & 38.90 & 55.47 \\
        \ \ \ + data-aug           & 82.53 & 77.03 & 93.70 & 85.36 \\
        \ \ \ + SW-SDRO    & \cellcolor{VeryLightGray}{83.92} & 75.82 & 88.92 & 81.48 \\ 
        \ \ \ + GW-SDRO     & \cellcolor{VeryLightGray}{84.05} & 76.95 & 93.41 & 85.18 \\
        \ \ \ \quad + Test-Time Ensembling     & \cellcolor{VeryLightGray}{84.22} & --''-- & --''-- & --''-- \\
        \midrule
        VILLA$_{BASE}$              & 84.82 & 74.15 & 37.40 & 55.77 \\
        \ \ \ + data-aug            & 83.54 & \textul{78.33} & \textul{94.55} & \textul{86.45} \\
        \ \ \ + SW-SDRO    & 84.54 & 74.02 & 88.32 & 81.17 \\
        \ \ \ + GW-SDRO     & \textbf{\cellcolor{VeryLightGray}{{85.12}}} & 77.92 & 93.42 & 85.67 \\
        \ \ \ \quad + Test-Time Ensembling     & \cellcolor{VeryLightGray}{\textbf{85.37}} & --''-- & --''-- & --''-- \\
        \bottomrule
    \end{tabular}
    }
    \caption[Results on the VQA yes/no subset.
    Not to be compared with VQA-v2 leaderboard since we use a smaller training set.]{Results on the VQA yes/no subset.\footnotemark[\value{footnote}] Not to be compared with VQA-v2 leaderboard since we use a smaller training set of \textit{yes/no} questions.
    % Note that these results should not be compared with VQA-v2 leaderboard since we use a smaller training set.
    }
    \label{tab:results_vqayesno}
\end{table}
We use UNITER and VILLA as the backbone models, with $T{=}20\%$.
The motivation behind VQA experiments is to show that SISP transforms and SDRO can be extended to other V\&L tasks.
Table~\ref{tab:results_vqayesno} shows that GW-SDRO is the best performing model in terms of clean accuracy, and is further improved by test-time ensembling.

%-------------------------------------------------------------------------
%%%%%%%%% ANALYSIS
\section{Analysis}
\label{sec:05_analysis}

\subsection{Visualization of Perturbations}
\begin{figure}
    \centering
    \includegraphics[width=\linewidth]{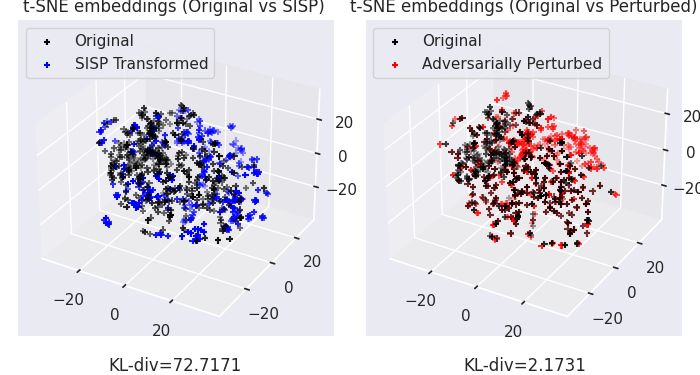}
    \caption{
    Comparison of original sentences (black) with \textit{(left)} SISP-transformed sentences (blue) and \textit{(right)} \straightepsilon-bounded perturbations
    as a tSNE plot.
    }
    \label{fig:tsne}
\end{figure}

In order to quantify the diverse and larger semantic transformations compared to additive perturbations, 
we study the tSNE~\cite{van2008visualizing} embeddings of (i) original samples from NLVR$^2$ ($P$), (ii) their SISP-transformed versions ($P_{SISP}$), and (iii) their adversarially perturbed versions ($P_{adv}$).
Input sentences are encoded using the UNITER text encoder for (i) and (ii), and the adversarial perturbation mechanism~\cite{gan2020large} for (iii).
3D tSNE embeddings are visualized in Figure~\ref{fig:tsne}; SISP transformed sentences (blue) are farther away than the perturbed versions.
This shift is quantified by the KL-divergence~\cite{kullback1951information} between the distributions, with $D_{KL}(P_{SISP}||P) > D_{KL}(P_{adv}||P)$ implying that the diversity of SISP transformations is higher.

\subsection{Comparison of Model Calibration}
\begin{figure*}
    \centering
    \includegraphics[width=\linewidth]{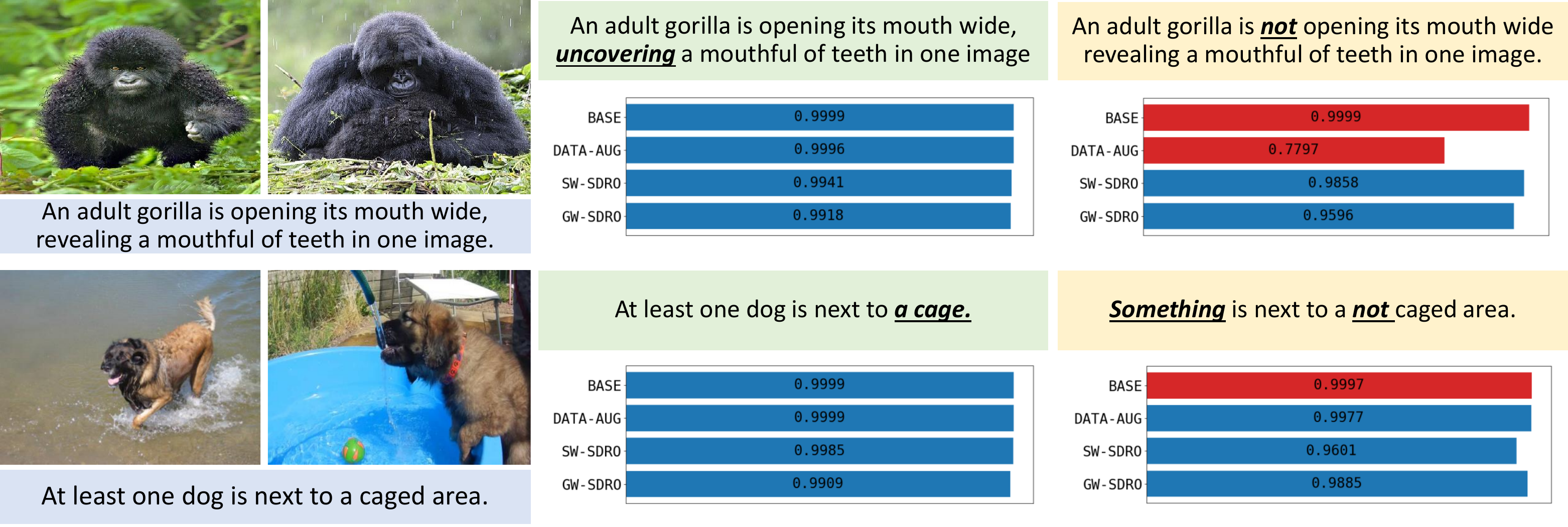}
    \caption{
    Qualitative examples showing test inputs from the NLVR$^2$ test set (left) with their respective SP (green) and SI (yellow) test samples.
    The predicted class (\texttt{True}/\texttt{False}) and the confidence of the predicted class is shown for baseline, data augmentation using SISP transforms, SW-SDRO and GW-SDRO.
    All models are built on the VILLA backbone.
    Wrong predictions are highlighted in \color{red}{red}.
    }
    \label{fig:qualitative}
\end{figure*}
Figure~\ref{fig:qualitative} contains qualitative examples from NLVR$^2$ to compare output probabilities.
We observe that SDRO models have higher clean accuracy, but lower confidence in the predictions than baseline and \textit{data-aug} methods.
\paragraph{Reliability Diagrams.}
\begin{figure}
    \centering
    \includegraphics[width=\linewidth]{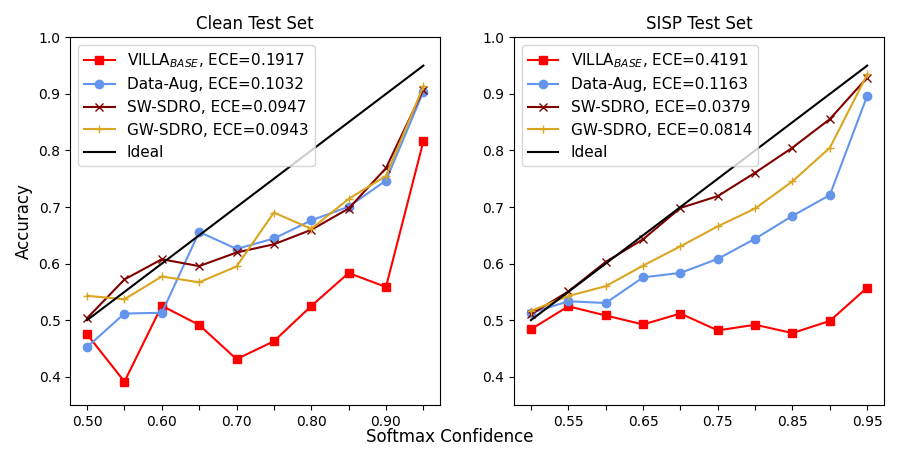}
    \caption{Comparison of reliability curves on the clean test set \textit{(left)} and SISP test set \textit{(right)}.}
    \label{fig:reliability}
\end{figure}
To validate this observation at scale, we use reliability diagrams to visualize model calibration~\cite{niculescu2005predicting}, and plot model accuracy as a function of confidence~\citep{guo2017calibration}.
We use the softmax probability $\hat{p}$ of the predicted class as model confidence, split the range of probabilities into $M=20$ equal-sized bins, and calculate bin accuracy $acc(B_m)$ and bin confidence $conf(B_m)$.
If $B_m$ is the set of all samples that fall in the $m^{th}$ bin,
\begin{align}
    \small
    \mathrm{acc}(B_m) &\triangleq \frac{1}{|B_m|}\sum_{X_i\in B_m}\mathbbm{1}(\hat{y}_i = y_i), \\
    \mathrm{conf}(B_m) &\triangleq \frac{1}{|B_m|}\sum_{X_i\in B_m} \hat{p}_i .
\end{align}
A model with perfect calibration should have a reliability diagram such that $acc(B_m){=}conf(B_m)$.
We also report Expected Calibration Error~\cite{naeini2015obtaining} over all $n$ test samples:
\begin{equation}
    ECE=\sum_{m=1}^M\frac{|B_m|}{n}|\mathrm{acc}(B_m){-}\mathrm{conf}(B_m)|.
\end{equation}

Reliability diagrams and corresponding ECE values for the VILLA trained with naive data augmentation and SDRO methods for NLVR$^2$ are shown in Figure~\ref{fig:reliability}.
On both the clean test set and SISP test set, SDRO models have the lowest ECE.
While the ECE for SDRO is marginally better than data augmentation for the clean test set, SDRO is better calibrated for the SISP test set, with SW-SDRO closest to ideal calibration among all evaluated models.

\subsection{Size of Training Dataset}
\begin{figure}[t]
    \centering
    \includegraphics[width=\linewidth]{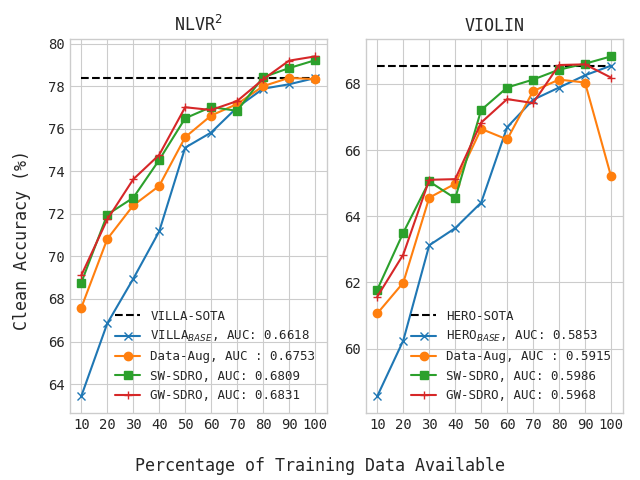}
    \caption{
    Effect of size of training data \textit{(left)} NLVR$^2$, \textit{(right)} VIOLIN.
    SDRO models are consistently better than baselines, even in low-data settings. 
    }
    \label{fig:lowres}
\end{figure}

We evaluate models trained on small subsets of the original dataset, and compare their performance in Figure~\ref{fig:lowres}.
SDRO models are significantly better at all sizes of training datasets as shown by accuracy and AUC (area under the curve).
Notably, SDRO models trained with only $10\%$ ($\sim8.6K$) samples have performances similar to the baseline trained with $30\%$ samples;
SDRO models with $20\%$ data are better than the baseline model with $40\%$ data.
While models trained with naive augmentation saturate below SOTA, at ${\sim}80\%$ data size, SDRO models cross the existing SOTA of $78.39\%$.

\begin{figure}[t]
    \centering
    \includegraphics[width=\linewidth]{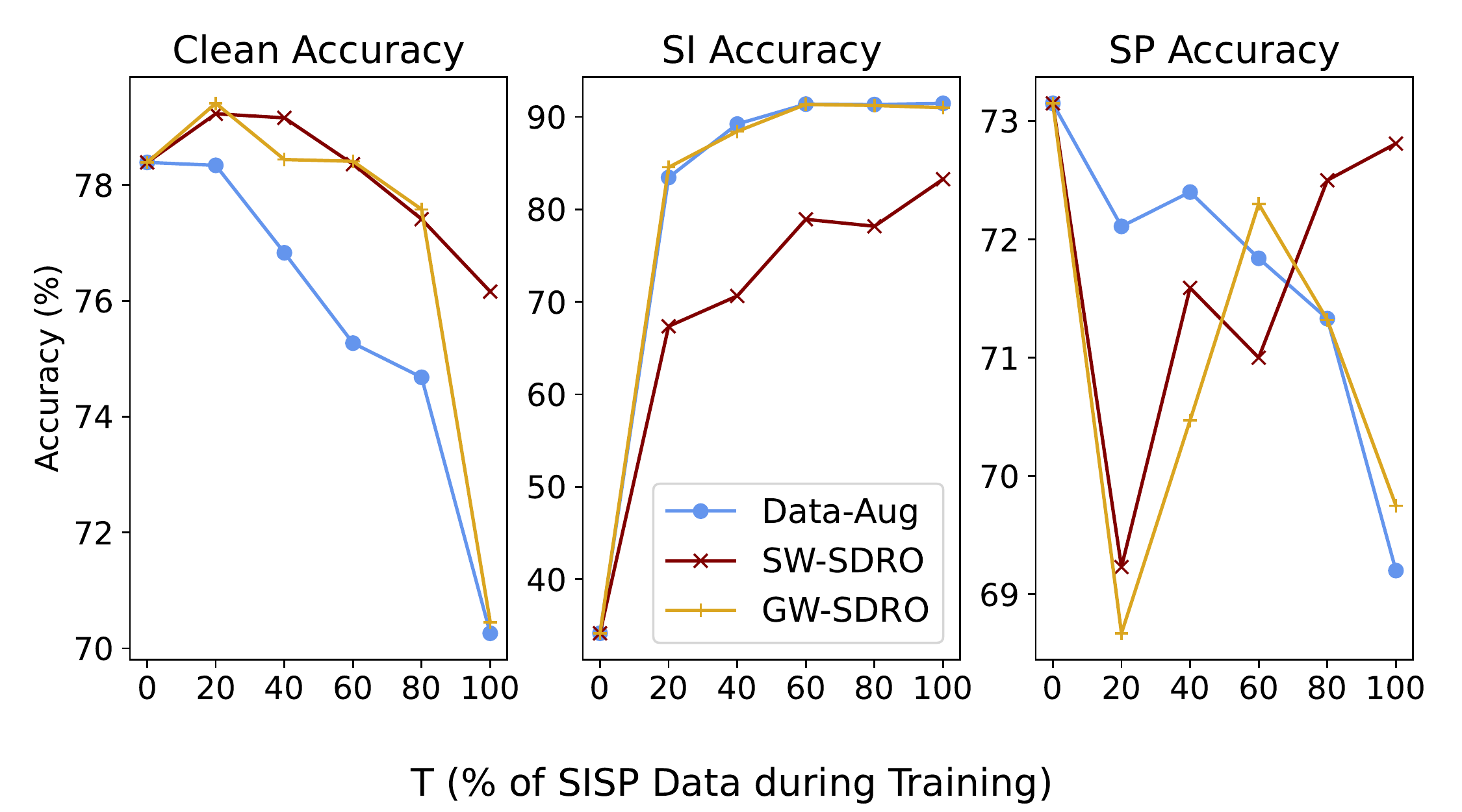}
    \caption{
    Plots showing the effect of the percentage of augmented samples on Clean, SP, and SI accuracies on NLVR$^2$, when using data-augmentation, and SDRO.
    }
    \label{fig:NLVR2_ablation_T}
\end{figure}
\begin{table*}[t]
    \centering
    \resizebox{0.68\linewidth}{!}{
    \begin{tabular}{@{}l ccc c ccc c ccc@{}}
        \toprule
        \multirow{2}{*}{\textbf{Model}} & \multicolumn{3}{c}{SP only} & \hphantom & \multicolumn{3}{c}{SI Only} & \hphantom & \multicolumn{3}{c}{Both} \\ 
         \cmidrule{2-4} \cmidrule{6-8}  \cmidrule{10-12}
         & Clean & SP & SI && Clean & SP & SI && Clean & SP & SI \\
        \midrule
        Data-Aug & 76.07 & 74.89 & 35.77 && 69.51 & 53.68 & 94.89 && 78.34 & 72.11 & 84.44 \\
        SW-SDRO & 79.79 & 76.93 & 30.72 && 79.27 & 55.53 & 88.76 && 79.23 & 69.23 & 67.35 \\
        GW-SDRO & 79.46 & 75.72 & 33.04 && 79.13 & 54.31 & 93.25 && 79.41 & 68.67 & 84.54 \\
        \bottomrule
    \end{tabular}
    }
    \caption{Comparison of performance when only SP, only SI, or both types of transformations are performed.}
    \label{tab:analysis_si_sp_all}
\end{table*} 
\begin{table}
    \centering
    % \footnotesize
    \resizebox{\linewidth}{!}{
    \begin{tabular}{@{}l ccc c ccc@{}}
        \toprule
        \multirow{2}{*}{\textbf{Model}} & \multicolumn{3}{c}{SISP~(Pos)} & \hphantom & \multicolumn{3}{c}{SISP~(All)} \\ 
         \cmidrule{2-4} \cmidrule{6-8}
         & Clean & SP & SI && Clean & SP & SI \\
        \midrule
        Data-Aug & 78.23 & 68.02 & 57.48 && 78.34 & 72.11 & 84.44 \\
        SW-SDRO  & 78.81 & 62.06 & 66.07 && 79.23 & 69.23 & 67.35 \\
        GW-SDRO  & 79.10 & 63.47 & 62.29 && 79.41 & 68.67 & 84.54\\
        \bottomrule
    \end{tabular}
    }
    \caption{Comparison of performance if only positive samples 
    % or both positive or negative samples 
    % i.e.\ samples with \texttt{True} labels 
    are used as inputs for SISP transformations
    % ,  transformations over both positive and negative samples.
    }
    \label{tab:analysis_pos_all}
\end{table} 

\subsection{Proportion of Augmented Samples.}
The final dataset has the same size as the original training set, but with T$\%$ transformed samples and $(100{-}T)\%$ original samples.
The effect of this hyperparameter $T$ is reported in Figure~\ref{fig:NLVR2_ablation_T} as a percentage improvement of accuracy w.r.t.\ VILLA$_{BASE}$.
An {optimal value of $T{=}20\%$ leads to improvements in clean accuracy}, but a larger proportion of augmented samples degrades performance.
Similarly, higher $T$ leads to higher robust accuracy, pointing to a \textul{trade-off between clean accuracy and robust accuracy at values of $T$ higher than the optimal}.
This conforms with similar findings from~\citet{tsipras2019robustness}.
While models trained with naive data-augmentation have better SISP accuracy than SDRO models as in Table~\ref{tab:results_nlvr2}, they do so by sacrificing clean accuracy, while SDRO models improve along both dimensions compared to the baselines.

\begin{table}[t]
    \centering
    % \small
    \resizebox{\linewidth}{!}{
    \begin{tabular}{@{}cl ccccccc@{}}
        \toprule
        & \textbf{Model} & \textbf{CR} & \textbf{CS} & \textbf{CL} & \textbf{EDA} & \textbf{Emb} & \textbf{WN} & \textbf{Avg.}\\
        \midrule
        \multirow{2}{*}{NLVR$^2$} 
        & VILLA          & 77.5 & 74.4 & 74.4 & 69.6 & 75.5 & 75.9 & 74.5 \\
        & ~~+ SDRO   & 78.5 & 77.2 & 72.1 & 71.1 & 75.8 & 76.4 & \textbf{75.2}\\
        \midrule
        \multirow{2}{*}{VIOLIN} 
        & HERO          & 66.1  & 63.0 & 68.6 & 60.9 & 63.8 & 63.4 & 64.3 \\
        & ~~+ SDRO   & 68.7 & 65.0 & 69.0 & 61.3 & 65.5 & 64.6 & \textbf{65.7} \\
        \midrule
        \multirow{2}{*}{\shortstack{VQA \\Yes/No}} 
        & VILLA          & 80.5 & 75.7 & 84.9 & 74.6 & 78.6 & 76.4 & 78.5 \\
        & ~~+ SDRO   & 86.0 & 84.5 & 84.1 & 87.0 & 84.3 & 84.0 & \textbf{85.0} \\
        \bottomrule
    \end{tabular}
    }
    \caption{Performance evaluation on ``text-attack''~\cite{morris2020textattack} versions of NLVR$^2$, VIOLIN, and VQA-Yes/No test sets. 
    % CR: CLARE, CS: CharSwap, CL: Check-List, EDA: Easy Data Aug, Emb: Embedding-based, WN: WordNet-based.
    }
    \label{tab:text_attack}
\end{table}
\subsection{Ablation Studies}
\paragraph{Contributions of SI and SP independently:}

We analyze which of the two categories (semantics-inverting (SI) or semantics-preserving (SP)) is the most effective by performing SDRO with only SI transforms, or with only SP transforms, and when using both.
Table~\ref{tab:analysis_si_sp_all} shows that SDRO models trained only with SI suffer in terms of SP robustness and vice versa.
However, there is still an increase in clean accuracy in both cases.
This indicates that \textul{both SI and SP contribute towards improvements in robustness and clean accuracy.}

\paragraph{Transformations of only \texttt{True} statements:}
Transforming \textit{False} (negative) statements can lead to ambiguous and subjective meanings~\cite{russell1905denoting}.
We investigate if transforming only \textit{True} (positive) statements is better than transforming both \textit{True} and \textit{False} statements.
Table~\ref{tab:analysis_pos_all} shows that SISP transformations of both types of statements lead to higher clean accuracy and robustness.

\subsection{Robustness to Text-Attacks}
In this section, we test each model against text-based adversarial attacks -- these attack samples are not seen by the models during training.
Thus, this experiment seeks to verify if training with SDRO and SISP samples can also make VLI models robust against automated adversarial attack recipes.
We utilize six common attack recipes implemented using the Text-Attack tool by \citet{morris2020textattack}; these are --
CLARE (CR)~\cite{li2021contextualized}, character-swap (CS)~\cite{pruthi2019combating}, Checklist (CL)~\cite{ribeiro2020beyond}, Easy Data Augmentation~\cite{wei2019eda}, counter-fitted embeddings (Emb.)~\cite{alzantot2018generating}, and WordNet-based swap (WN)~\cite{ren2019generating}.
Table~\ref{tab:text_attack} shows results on each benchmark, using the best performing backbone for that benchmark and our SDRO model.
On NLVR$^2$, VILLA+SDRO is better than VILLA for 4 out of 6 attack categories, and $0.7\%$ on average.
On VIOLIN, HERO+SDRO outperforms the baseline on all attack categories, leading to an average gain of $1.4\%$.
On VQA-Yes/No, VILLA+SDRO outperforms the baseline on all attack categories, and $6.5\%$ on average.

%-------------------------------------------------------------------------
%%%%%%%%% RELATED WORK
\section{Related Work}

\textbf{Adversarial Training} (AT)
has been studied under a game-theoretic~\cite{dalvi2004adversarial} and
min-max setup~\cite{madry2018towards}.
\citet{volpi2018generalizing} use AT to adversarially augment image classification datasets and show improved domain generalization for digit classification.
\citet{wong2020learning,gokhale2021attribute} modify AT for real-world adversaries beyond norm-bounded perturbations.
AT has been used for text classification with LSTMs~\cite{miyato2016adversarial} and for pretraining transformer-based models by adding label-preserving adversarial perturbations to embeddings of word tokens~\cite{zhu2020freelb,jiang2020smart,gan2020large}
Contrastive examples have been explored, collected from humans~\cite{agrawal2018don}, negative mining~\cite{shi2018learning}, or synthetic generation~\cite{agarwal2020towards,chen2020counterfactual,gokhale2020mutant,teney2020learning}.

\textbf{Robustness in V\&L}
has been explored for VQA, such as performance under prior probability shift~\cite{agrawal2018don} and domain adaptation~\cite{chao2018cross,xu2019open}, along with robustness for implied questions~\cite{ribeiro2019red} and novel compositions~\cite{johnson2017clevr,agrawal2017c}, and robustness to logical connectives (including negation)~\citet{gokhale2020vqa}.
\citet{teney2020learning} have shown that many V\&L, image classification, and sentiment analysis models are sensitive to image editing.
There has been a recent effort of model-in-the-loop dataset collection to guide humans to create harder VQA samples~\cite{li2021adversarial,sheng2021human}.

\textbf{Robustness in NLP:}
Generation of SP adversarial examples 
\cite{jia2017adversarial,ribeiro2018semantically,iyyer2018adversarial,alzantot2018generating}, and approaches to defend against word substitution~\cite{jia2019certified} have been explored in natural language processing tasks.
Evaluation datasets have also been proposed for textual entailment that are manually crafted~\cite{gardner2020evaluating} or template-based~\cite{mccoy2019right,glockner2018breaking,naik2018stress}.
Our method uses automated linguistically-informed SI and SP transforms for both training and inference.
%-------------------------------------------------------------------------
%%%%%%%%% DISCUSSION 
\section{Discussion}
\paragraph{On Ensembling Coefficients.}
While designing our ensembling approach, we used $\alpha=0.5$, i.e., equal contribution from the original output and the average of all outputs for transformed samples.
This choice is generic and does not rely on dataset- or model-specific characteristics of SISP accuracy.
While treating $\alpha$ as a hyperparameter and tuning it on validation datasets could lead to further gains, our intuitive choice of $\alpha=0.5$ is effective by itself.

\paragraph{On SI Samples.}
Tables~\ref{tab:results_nlvr2},~\ref{tab:results_violin},~\ref{tab:results_vqayesno} show that existing models perform well on SP transforms, implying that equivalent semantics are captured in transformer-based models.
However, these models fail on SI samples resulting in a close-to-random ($50\%$) average SISP accuracy.
While images perturbed with noise, blur, weather, or digital artifacts~\cite{hendrycks2018benchmarking} retain semantics (an image of a ``cat'' remains a cat after perturbation), minimal changes to text inputs, such as a single word changing from ``sitting'' to ``standing'' or ``not sitting'', inflict large changes in meaning.
We hope that future work on design of V\&L evaluation criterion along the SI axis, could benefit from our findings.
While we generated SI and SP text for VLI tasks, the idea could be extended to design SISP transformations for images, by operating at object-level instead of pixel-level

\paragraph{On combination of AT and SDRO.}
We show that combining AT with SDRO can improve VLI performance and incorporate domain knowledge into the training process, such as semantic knowledge that often exists in natural language or linguistic rules.
This is explicitly observed with VILLA, which is pre-trained and fine-tuned using standard adversarial training~\cite{gan2020large}.
When fine-tuned with SDRO, VILLA+SDRO further improves compared to UNITER+SDRO.
The combination of standard adversarial training, (which accounts for local adversaries inside a $\epsilon$ norm-ball) and SDRO, (which accounts for linguistic adversaries and contrastive examples, typically outside the norm-ball as shown in Figure~\ref{fig:vector_vs_linguistic}) could lead to improved generalization in many other V\&L tasks.

\paragraph{On differentiability.}
Linguistic transformations are not differentiable and prohibit gradient-based solutions to the inner maximization in SDRO.
However, most V\&L tasks would benefit from the incorporation of semantic knowledge into the optimization framework.
Through SDRO, we show that explicitly choosing the $argmax$ over a pre-defined set of transformations leads to model-agnostic improvements for binary classification tasks in V\&L.
More sophisticated methods may emerge in the future to address non-differentiability by leveraging proximal point or trust-region methods~\cite{eckstein1993nonlinear,conn2000trust} or Interval Bound Propagation~\cite{dvijotham2018dual}, to incorporate semantic knowledge into adversarial training.

%-------------------------------------------------------------------------
\section*{Acknowledgements}
This work was funded in part by National Science Foundation grants 2132724, 1816039 and 1750082, DARPA SAIL-ON program (W911NF2020006), and DARPA CHESS program (FA875019C0003). 
The authors are grateful to the volunteers who worked on rating the fidelity of our proposed transformations.
The views and opinions of the authors expressed herein do not necessarily state or reflect those of the funding agencies and employers.
\bibliographystyle{acl_natbib}
\bibliography{egbib}

\section*{Appendix}
\appendix
\noindent In this supplementary material, we provide finegrained results of our experiments, along with detailed analysis for VIOLIN and VQA-Yes/No similar to Section 5 in the main paper.
We also provide visualizations of the SISP data creation process, statistics for SISP-transformed samples, and details of our human evaluation study.

\section{Fine-Grained Results}

\subsection{Baseline Performance on SISP}
In Tables~\ref{tab:nlvr2_baseline_finegrained}, ~\ref{tab:violin_baseline_finegrained}, ~\ref{tab:vqa_baseline_finegrained} we compare the performance of baseline models on all 13 categories of SISP transforms.
All baseline models are below random performance on all three datasets for all SI categories, except for VIOLIN$_{BASE}$~\cite{liu2020violin}.
This is an interesting finding since VIOLIN$_{BASE}$ is the only model that is not a pretrained transformer-based model, but uses simple fusion of visual and textual modalities.
In this paper, we've considered 3 benchmarks, and $3+2+3=8$ backbone models in total.
Of these, only VIOLIN$_{BASE}$-- a non-transformer model, retains above-random performance on SISP samples.
Performance on SP categories is the best for VILLA~\cite{gan2020large} for NLVR$^2$ and VQA Yes/No, and HERO~\cite{li2020hero} for VIOLIN. 
\begin{table}[t]
    \centering
    % \footnotesize
    \resizebox{\linewidth}{!}{
    \begin{tabular}{@{}ll c c c@{}}
        \toprule
        & \textbf{Category} & \textbf{LXMERT} & \textbf{UNITER} &\textbf{VILLA}\\
        \toprule
        & Original & 74.37 & 77.85 & 78.39  \\
        \midrule
        \multirow{6}{*}{\rotatebox{90}{\footnotesize SI }} 
        & Comparative Antonym   & 49.19 & 40.11 & 34.32 \\
        & Negation   & 35.19 & 36.92 & 35.39 \\
        & Noun Antonym   & 29.94 & 35.35 & 39.05 \\
        & Number Substitution   & 45.26 & 39.53 & 35.24 \\
        & Pronoun Substitution   & 47.76 & 34.79 & 29.78 \\
        & Subject-Object Swap   & 20.26 & 27.65 & 30.41 \\
        & Verb Antonym   & 27.86 & 29.72 & 34.89 \\
        \midrule
        \multirow{6}{*}{\rotatebox{90}{\footnotesize SP}} &
        Comparative Synonym   & 61.35 & 65.58 & 66.86 \\
        & Paraphrasing   & 71.33 & 73.62 & 73.46 \\
        & Noun Synonym   & 71.24 & 75.32 & 75.78 \\
        & Number Substitution   & 70.68 & 74.33 & 74.37 \\
        & Pronoun Substitution   & 69.36 & 73.36 & 73.16 \\
        & Verb Synonym   & 71.26 & 74.16 & 75.24 \\
        \bottomrule 
    \end{tabular}
    }
    \caption{Evaluation of NLVR2 baselines on SISP test samples.}
    \label{tab:nlvr2_baseline_finegrained}
\end{table}
\begin{table}[ht]
    \centering
    % \footnotesize
    \resizebox{\linewidth}{!}{
    \begin{tabular}{@{}ll c c c@{}}
        \toprule
        & \textbf{Category} & \textbf{VIOLIN$_{BASE}$} & \textbf{HERO$_{BASE}$} \\
        \toprule
        & Original & 68.07 & 68.55\\
        \midrule
        \multirow{6}{*}{\rotatebox{90}{\footnotesize SI }} 
        & Comparative Antonym   & 58.33 & 31.66 \\
        & Negation   & 57.75 & 34.73 \\
        & Noun Antonym   & 57.21 & 37.06 \\
        & Number Substitution   & 54.21 & 26.07 \\
        & Pronoun Substitution   & 57.66 & 24.64 \\
        & Subject-Object Swap   & 57.59 & 31.13 \\
        & Verb Antonym   & 57.68 & 38.77 \\
        \midrule
        \multirow{6}{*}{\rotatebox{90}{\footnotesize SP}} &
        Comparative Synonym   & 57.92 & 67.87 \\
        & Paraphrasing   & 57.32 & 65.81 \\
        & Noun Synonym   & 57.67 & 67.15 \\
        & Number Substitution   & 54.87 & 58.88 \\
        & Pronoun Substitution   & 57.68 & 66.74 \\
        & Verb Synonym   & 57.53 &  67.09 \\
        \bottomrule 
    \end{tabular}
    }
    \caption{Evaluation of VIOLIN baselines on SISP test samples.}
    \label{tab:violin_baseline_finegrained}
\end{table}
\begin{table}[ht]
    \centering
    % \footnotesize
    \resizebox{\linewidth}{!}{
    \begin{tabular}{@{}ll c c c@{}}
        \toprule
        & \textbf{Category} & \textbf{LXMERT} & \textbf{UNITER} &\textbf{VILLA}\\
        \toprule
        & Original & 83.13 & 83.655 & 84.82  \\
        \midrule
        \multirow{6}{*}{\rotatebox{90}{\footnotesize SI }} 
        & Comparative Antonym   & 36.7 & 39.07 & 39.59 \\
        & Negation   & 29.59 & 31.93 & 29.59 \\
        & Noun Antonym   & 48.36 & 53.21 & 50.88 \\
        & Number Substitution   & 26.32 & 42.11 & 49.47 \\
        & Pronoun Substitution   & 21.28 & 24.05 & 24.36 \\
        & Subject-Object Swap   & 24.68 & 31.33 & 26.06 \\
        & Verb Antonym   & 35.88 & 50.63 & 41.86 \\
        \midrule
        \multirow{6}{*}{\rotatebox{90}{\footnotesize SP}} &
        Comparative Synonym   & 67.72 & 71.28 & 74.11 \\
        & Paraphrasing   & 79.63 & 79.37 & 80.74 \\
        & Noun Synonym   & 74.09 & 73.37 & 74.61 \\
        & Number Substitution   & 72.32 & 57.89 & 62.11 \\
        & Pronoun Substitution   & 74.82 & 76.11 & 77.48 \\
        & Verb Synonym   & 73.76 & 74.22 & 75.82 \\
        \bottomrule 
    \end{tabular}
    }
    \caption{Evaluation of VQA Yes/No baselines on SISP test samples.}
    \label{tab:vqa_baseline_finegrained}
\end{table}

\subsection{SDRO Performance on SISP}
In Tables~\ref{tab:nlvr2_sisp_dataset_sota_eval},~\ref{tab:violin_sisp_dataset_sota_eval},~\ref{tab:vqa_sisp_dataset_sota_eval} we compare performance for the state-of-the-art model VILLA, as well as models trained with naive data augmentation and our SDRO methods.

\begin{table}[t]
    \centering
    % \footnotesize
    \resizebox{\linewidth}{!}{
    \begin{tabular}{@{}l l c c c c@{}}
        \toprule
        & \textbf{Category} & \textbf{BASE} & \textbf{Data-Aug} &\textbf{SW-SDRO} & \textbf{GW-SDRO}\\
        \toprule
        & Original & 78.39 & 78.34 & 79.23 &  79.41 \\
        \midrule
        \multirow{6}{*}{\rotatebox{90}{\footnotesize SI }} 
        & Noun Antonym   & 39.05 & 85.79 & 63.13 & 76.64 \\
        & Negation   & 35.39 & 65.75 & 72.78 & 57.29 \\
        & Subject-Object Swap   & 30.41 & 87.13 & 60.19 & 89.06 \\
        & Verb Antonym  & 34.89 & 72.58 & 55.18 & 84.19 \\
        & Number Substitution    & 35.24 & 95.79 & 75.79 & 93.07 \\
        & Pronoun Substitution   & 29.78 & 98.44 & 81.31 & 98.35 \\
        & Comparative Antonym   & 34.32 & 78.62 & 63.11 & 93.17 \\
        \midrule
        \multirow{6}{*}{\rotatebox{90}{\footnotesize SP}} &
        Pronoun Substitution   & 73.16 & 72.81 & 64.91 & 69.68\\
        & Number Substitution   & 74.37 & 81.27 & 77.63 & 78.42 \\
        & Comparative Synonym   & 66.88 & 64.63 & 64.16 & 66.32 \\ 
        & Verb Synonym   & 75.24 & 69.88 & 65.78 & 59.83 \\
        & Paraphrasing   & 73.46 & 74.89 & 75.74 & 76.13 \\
        & Noun Synonym   & 75.78 & 69.15 & 67.67 & 61.64\\
        \bottomrule 
    \end{tabular}
    }
    \caption{Evaluation of SDRO models (with VILLA bacbone) on  NLVR$^2$ SISP test samples.}
    \label{tab:nlvr2_sisp_dataset_sota_eval}
\end{table}

\begin{table}[t]
    \centering
    % \footnotesize
    \resizebox{\linewidth}{!}{
    \begin{tabular}{@{}l l c c c c@{}}
        \toprule
        & \textbf{Category} & \textbf{BASE} & \textbf{Data-Aug} &\textbf{SW-SDRO} & \textbf{GW-SDRO}\\
        \toprule
        & Original & 68.55 & 65.21 & 68.83 & 68.19  \\
        \midrule
        \multirow{6}{*}{\rotatebox{90}{\footnotesize SI }} 
        & Noun Antonym   & 37.06 & 86.38 & 76.61 & 95.19 \\
        & Negation   & 34.73 & 53.18 & 58.31 & 61.14 \\
        & Subject-Object Swap   & 31.13 & 94.28 & 74.98 & 95.08 \\
        & Verb Antonym  & 38.77 & 81.96 & 77.05 & 94.95 \\
        & Number Substitution    & 26.07 & 76.32 & 80.03 & 71.04 \\
        & Pronoun Substitution   & 24.64 & 99.41 & 92.89 & 98.03 \\
        & Comparative Antonym   & 31.66 & 81.12 & 84.44 & 65.04 \\
        \midrule
        \multirow{6}{*}{\rotatebox{90}{\footnotesize SP}} &
        Pronoun Substitution   & 66.74 & 62.29 & 60.76 & 69.67\\
        & Number Substitution   & 58.88 & 56.62 & 57.14 & 51.22 \\
        & Comparative Synonym   & 67.87 & 58.31 & 57.64 & 49.67 \\ 
        & Verb Synonym   & 67.09 & 56.42 & 56.49 & 50.15 \\
        & Paraphrasing   & 65.81 & 63.59 & 65.22 & 66.03 \\
        & Noun Synonym   & 67.15 & 57.99 & 56.67 & 50.47\\
        \bottomrule 
    \end{tabular}
    }
    \caption{Evaluation of SDRO models (with HERO backbone) on VIOLIN SISP Data}
    \label{tab:violin_sisp_dataset_sota_eval}
\end{table}

\begin{table}[t]
    \centering
    % \footnotesize
    \resizebox{\linewidth}{!}{
    \begin{tabular}{@{}l l c c c c@{}}
        \toprule
        & \textbf{Category} & \textbf{BASE} & \textbf{Dataaug} &\textbf{SW-SDRO} & \textbf{GW-SDRO}\\
        \toprule
        & Original & 84.82 & 83.54 & 84.88 &  85.19 \\
        \midrule
        \multirow{6}{*}{\rotatebox{90}{\footnotesize SI }} 
        & Noun Antonym   & 50.88 & 97.85 & 92.04 & 92.06 \\
        & Negation   & 29.59 & 80.81 & 82.36 & 81.39 \\
        & Subject-Object Swap   & 26.06 & 98.83 & 96.19 & 98.98 \\
        & Verb Antonym  & 41.86 & 97.71 & 88.17 & 98.6 \\
        & Number Substitution    & 49.47 & 94.74 & 78.95 & 92.63 \\
        & Pronoun Substitution   & 24.36 & 95.36 & 90.86 & 94.24 \\
        & Comparative Antonym   & 39.59 & 96.58 & 89.64 & 96.05 \\
        \midrule
        \multirow{6}{*}{\rotatebox{90}{\footnotesize SP}} &
        Pronoun Substitution   & 77.48 & 77.41 & 75.88 & 76.98 \\
        & Number Substitution   & 62.11 & 77.89 & 56.84 & 81.05 \\
        & Comparative Synonym   & 74.11 & 80.72 & 78.63 & 80.25 \\ 
        & Verb Synonym   & 75.82 & 76.85 & 76.13 & 75.51 \\
        & Paraphrasing   & 80.74 & 80.57 & 81.31 & 81.49 \\
        & Noun Synonym   & 74.61 & 76.55 & 75.27 & 72.23\\
        \bottomrule 
    \end{tabular}
    }
    \caption{Evaluation of SDRO models (with VILLA backbone) on VQA Yes/No  SISP Data}
    \label{tab:vqa_sisp_dataset_sota_eval}
\end{table}

%%%%%%%%%%%%%%%%%%%%%%%%%%%%%%%%
\begin{figure}[t]
    \centering
    \includegraphics[width=\linewidth]{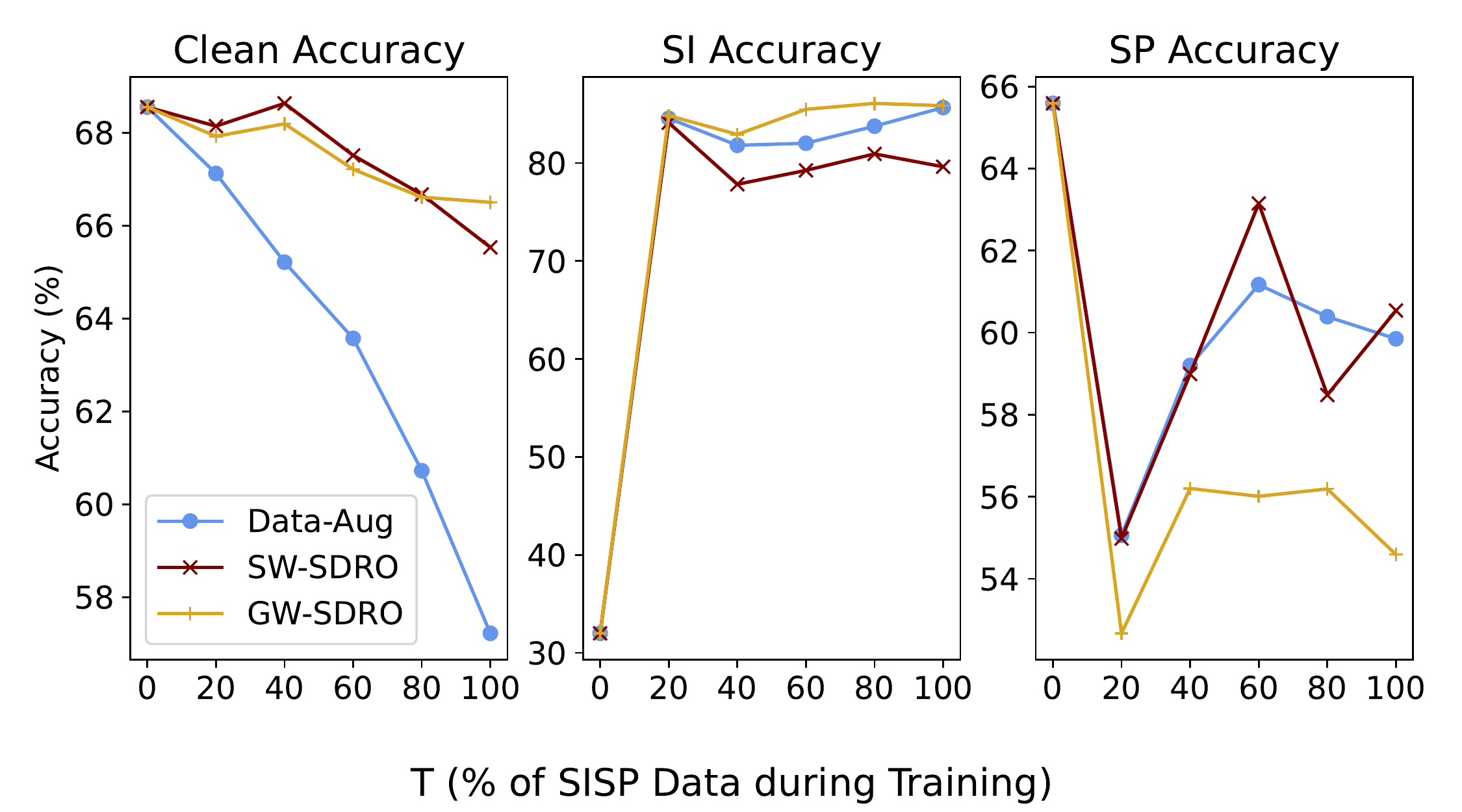}
    \caption{Plots showing the effect of the percentage of augmented samples on Clean, SP, and SI accuracies on VIOLIN, when using naive data-augmentation, SW-SDRO, and GW-SDRO.
    }
    \label{fig:violin_ablation_T}
\end{figure}
\section{Analysis for VIOLIN}
\paragraph{Proportion of Augmented Samples.}
We perform an analysis by varying $T$ (proportion of augmented samples) and report performance in Figure~\ref{fig:violin_ablation_T} as a percentage improvement of accuracy w.r.t.\ HERO$_{BASE}$.
It can be seen that there exists an \textul{optimal value of $T$ ($40\%$), which leads to improvements in clean accuracy}, but higher values of T, i.e., a larger proportion of augmented samples degrades performance.
Similarly, higher $T$ leads to higher robust accuracy, but lower clean accuracy.
Models trained with naive data-augmentation may be more robust on SP and SI test samples than SDRO models, but they do so by sacrificing clean accuracy, while SDRO models improve along both dimensions compared to the baselines.

\paragraph{Contributions of SI and SP independently}
Table~\ref{tab:violin_si_sp_all} shows that SDRO models trained only with SI suffer in terms of SP robustness and vice versa.
However, there is still an increase in clean accuracy in both cases, thus indicating the efficacy of both SP and SI transformations.
\begin{table*}[ht]
    \centering
    \footnotesize
    % \resizebox{\linewidth}{!}{
    \begin{tabular}{@{}l ccc c ccc c ccc@{}}
        \toprule
        \multirow{2}{*}{\textbf{Model}} & \multicolumn{3}{c}{SP only} & \hphantom & \multicolumn{3}{c}{SI Only} & \hphantom & \multicolumn{3}{c}{Both} \\ 
         \cmidrule{2-4} \cmidrule{6-8}  \cmidrule{10-12}
         & Clean & SP & SI && Clean & SP & SI && Clean & SP & SI \\
        \midrule
        Data-Aug    & 63.33 & 60.39 & 31.71 && 66.11 & 50.28 & 87.29 && 65.21 & 59.20 & 81.81\\
        SW-SDRO     & 67.18 & 65.49 & 31.49 && 67.11 & 50.64 & 85.16 && 68.83 & 58.97 & 77.83 \\
        GW-SDRO     & 67.73 & 65.93 & 30.80 && 67.43 & 51.21 & 87.72 && 68.19 & 56.20 & 82.92 \\
        \bottomrule
    \end{tabular}
    % }
    \caption{Comparison of performance on the VIOLIN dataset when only SP, only SI, or both types of transformations are performed.}
    \label{tab:violin_si_sp_all}
\end{table*}
\begin{table*}
    \centering
    \footnotesize
    % \resizebox{\linewidth}{!}{
    \begin{tabular}{@{}l ccc c ccc c ccc@{}}
        \toprule
        \multirow{2}{*}{\textbf{Model}} & \multicolumn{3}{c}{SP only} & \hphantom & \multicolumn{3}{c}{SI Only} & \hphantom & \multicolumn{3}{c}{Both} \\ 
         \cmidrule{2-4} \cmidrule{6-8}  \cmidrule{10-12}
         & Clean & SP & SI && Clean & SP & SI && Clean & SP & SI \\
        \midrule
        Data-Aug & 82 & 78.73 & 31.91 && 84.2 & 54.88 & 95.5 && 83.54 & 78.33 & 94.55 \\
        SW-SDRO & 84.01 & 79.46 & 33.28 && 84.23 & 52.59 & 94.28 && 84.88 & 74.02 & 88.32 \\
        GW-SDRO & 85.03 & 79.31 & 32.57 && 85.01 & 53.46 & 95.61 && 85.19 & 77.92 & 93.42 \\
        \bottomrule
    \end{tabular}
    % }
    \caption{Comparison of performance on the VQA Yes/No dataset when only SP, only SI, or both transformations are performed.}
    \label{tab:vqa_si_sp_all}
\end{table*}
\begin{table}[ht]
    \centering
    % \footnotesize
    \resizebox{\linewidth}{!}{
    \begin{tabular}{@{}l ccc c ccc@{}}
        \toprule
        \multirow{2}{*}{\textbf{Model}} & \multicolumn{3}{c}{$\mathbf{SISP(Pos)}$} & \hphantom & \multicolumn{3}{c}{$\mathbf{SISP(All)}$} \\ 
         \cmidrule{2-4} \cmidrule{6-8}
         & Clean & SP & SI && Clean & SP & SI \\
        \midrule
        Data-Aug & 66.22 & 49.16 & 82.66 && 65.21 & 59.20 & 81.81  \\
        SW-SDRO  & 67.99 & 56.08 & 79.06 && 68.83 & 58.97 & 77.83 \\
        GW-SDRO  & 67.34 & 55.19 & 82.90 && 68.19 & 56.20 & 82.92 \\
        \bottomrule
    \end{tabular}
    }
    \caption{Comparison of performance on VIOLIN dataset if only positive samples, i.e.\ samples with \texttt{True} labels are used as inputs for SISP transformations.
    % , vs. transformations over both positive and negative samples.
    }
    \label{tab:violin_pos_all}
\end{table}
\begin{table}[ht]
    \centering
    % \footnotesize
    \resizebox{\linewidth}{!}{
    \begin{tabular}{@{}l ccc c ccc@{}}
        \toprule
        \multirow{2}{*}{\textbf{Model}} & \multicolumn{3}{c}{$\mathbf{SISP(Pos)}$} & \hphantom & \multicolumn{3}{c}{$\mathbf{SISP(All)}$} \\ 
         \cmidrule{2-4} \cmidrule{6-8}
         & Clean & SP & SI && Clean & SP & SI \\
        \midrule
        Data-Aug & 84.2 & 64.48 & 60.36 && 83.54 & 78.33 & 94.55 \\
        SW-SDRO  & 84.73 & 61.13 & 61.64 && 84.88 & 74.02 & 88.32 \\
        GW-SDRO  & 84.99 & 62.65 & 62.44 && 85.19 & 77.92 & 93.42 \\
        \bottomrule
    \end{tabular}
    }
    \caption{Comparison of performance on VQA-Yes/No dataset if only positive samples, i.e.\ samples with \texttt{True} labels are used as inputs for SISP transformations, vs. transformations over both positive and negative samples.}
    \label{tab:vqa_pos_all}
\end{table}

\paragraph{Transformations of only \texttt{True} statements}
Table~\ref{tab:violin_pos_all} shows that training with transformations of both \texttt{True} and \texttt{False} helps both robustness and accuracy.

%%%%%%%%%%%%%%%%%%%%%%%%%%%%%%%%

\section{Analysis for VQA-Yes/No}
\paragraph{Proportion of Augmented Samples.}
\begin{figure}[t]
    \centering
    \includegraphics[width=\linewidth]{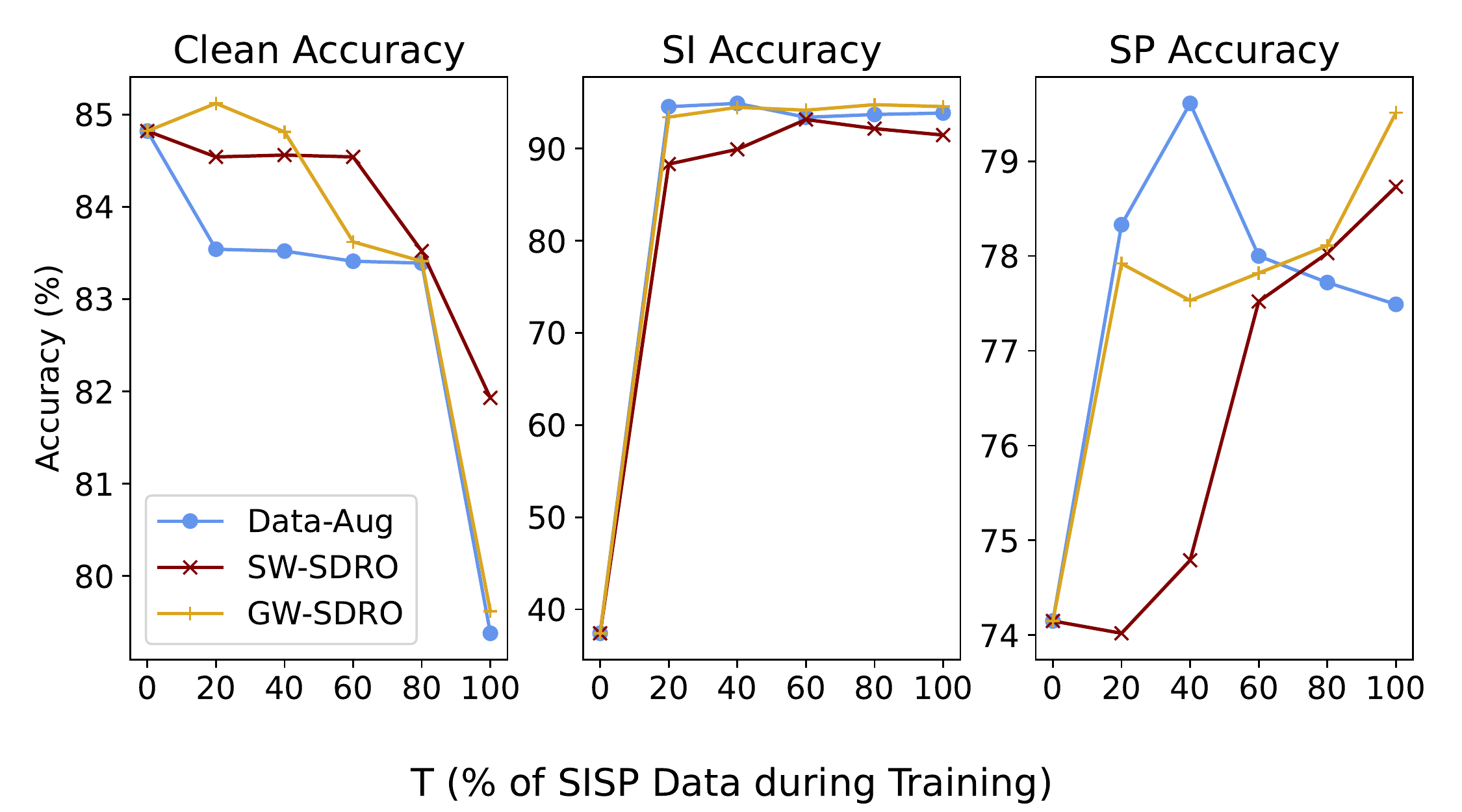}
    \caption{Plots showing the effect of the percentage of augmented samples on Clean, SP, and SI accuracies on VQA Yes/No, when using naive data-augmentation, SW-SDRO, and GW-SDRO.
    }
    \label{fig:vqa_ablation_T}
\end{figure}
We perform an analysis by varying $T$ (proportion of augmented samples) and report performance in Figure~\ref{fig:vqa_ablation_T} as a percentage improvement of accuracy w.r.t.\ VILLA$_{BASE}$.
Higher $T$ leads to higher robust accuracy, but lower clean accuracy.

\paragraph{Contributions of SI and SP independently}
Table~\ref{tab:vqa_si_sp_all} shows that SDRO models trained only with SI suffer in terms of SP robustness and vice versa.
However, there is still an increase in clean accuracy in both cases, thus indicating the efficacy of both SP and SI transformations.
The increase in clean accuracy is greater for models trained with both SP and SI transformations.

\paragraph{Transformations of only \texttt{True} statements}
Table~\ref{tab:vqa_pos_all} shows that training with transformations of both \texttt{True} and \texttt{False} helps both robustness and accuracy.

\section{SISP Dataset}
\subsection{Statistics}
\begin{table}[ht]
    \centering
    % \small
    \resizebox{\linewidth}{!}{
    \begin{tabular}{@{}ll c c c@{}}
        \toprule
        & \textbf{Category} & \textbf{Training} & \textbf{Test-P} &\textbf{Validation}\\
        \toprule
        & Original & 86,373 & 6,967 & 6,982  \\
        \midrule
        \multirow{6}{*}{\rotatebox{90}{\footnotesize SI }} 
        & Comparative Antonym   & 14,177 & 1,244 & 1,172 \\
        & Negation   & 150,610 & 12,838 & 12,635 \\
        & Noun Antonym   & 148,959 & 12,719 & 12,635 \\
        & Number Substitution   & 83,080 & 7,468 & 7,113 \\
        & Pronoun Substitution   & 34,145 & 3,210 & 2,997 \\
        & Subject-Object Swap   & 30,533 & 2,944 & 2,787 \\
        & Verb Antonym   & 24,711 & 2,258 & 2,258 \\
        \midrule
        & Total SI & 486215 & 42681 & 41714 \\
        \midrule
        \multirow{6}{*}{\rotatebox{90}{\footnotesize SP}} &
        Comparative Synonym   & 13,302 & 1,066 & 1,163 \\
        & Paraphrasing   & 86,373 & 6,967 & 6,982 \\
        & Noun Synonym   & 212,904 & 18,570 & 18,968 \\
        & Number Substitution   & 60,582 & 5,194 & 4,994 \\
        & Pronoun Substitution   & 32,508 & 2,869 & 2,852 \\
        & Verb Synonym   & 78,103 & 7,314 & 6,919 \\
        \midrule
        & Total SP & 483772 & 41980 & 41878 \\
        \bottomrule 
    \end{tabular}
    }
    \caption{Number of SISP-transformed samples generated per category for the NLVR2 dataset.}
    \label{tab:nlvr2_sisp_dataset_split}
\end{table}
\begin{table}[ht]
    \centering
    \footnotesize
    \resizebox{\linewidth}{!}{
    \begin{tabular}{@{}ll c c c@{}}
        \toprule
        & \textbf{Category} & \textbf{Training} & \textbf{Testing} &\textbf{Validation}\\
        \toprule
        & Original & 76122 & 9600 & 9600  \\
        \midrule
        \multirow{6}{*}{\rotatebox{90}{\footnotesize SI }} 
        & Comparative Antonym   & 66,300 & 8,754 & 8,893 \\
        & Negation   & 249,836 & 31,634 & 31,923 \\
        & Noun Antonym   & 193,964 & 24,484 & 24,251 \\
        & Number Substitution   & 9,592 & 1,212 & 1,130 \\
        & Pronoun Substitution   & 156,466 & 19,785 & 20,100 \\
        & Subject-Object Swap   & 122,510 & 15,500 & 15,337 \\
        & Verb Antonym   & 49,802 & 6,358 & 6,356 \\
        \midrule
        & Total SI & 848470 & 107727 & 107990 \\
        \midrule
        \multirow{6}{*}{\rotatebox{90}{\footnotesize SP}} &
        Comparative Synonym   & 38,312 & 4,955 & 4,940 \\
        & Paraphrasing   & 76,122 & 9,600 & 9600 \\
        & Noun Synonym   & 418,285 & 52,857 & 52002 \\
        & Number Substitution   & 4,482 & 574 & 544 \\
        & Pronoun Substitution   & 91,125 & 11,464 & 11539 \\
        & Verb Synonym   & 196,826 & 25,044 & 25576 \\
        \midrule
        & Total SP & 825152 & 104494 & 104201 \\
        \bottomrule 
    \end{tabular}
    }
    \caption{Number of SISP-transformed samples generated per category for the VIOLIN dataset.}
    \label{tab:violin_sisp_dataset_split}
\end{table}
\begin{table}[ht]
    \centering
    % \footnotesize
    \resizebox{\linewidth}{!}{
    \begin{tabular}{@{}ll c c c@{}}
        \toprule
        & \textbf{Category} & \textbf{Train} & \textbf{Trainval} &\textbf{Devval}\\
        \toprule
        & Original & 92,761 & 38,374 & 5,323  \\
        \midrule
        \multirow{6}{*}{\rotatebox{90}{\footnotesize SI }} 
        & Comparative Antonym   & 18,839 & 8,044 & 1,139 \\
        & Negation   & 100,302 & 41,676 & 5,738 \\
        & Noun Antonym   & 82,885 & 34,543 & 4,835 \\
        & Number Substitution   & 1,505 & 730 & 95 \\
        & Pronoun Substitution   & 26,804 & 11,462 & 1,597 \\
        & Subject-Object Swap   & 11,793 & 4,999 & 683 \\
        & Verb Antonym   & 13,262 & 5,707 & 786 \\
        \midrule
        & Total SI & 255390 & 107161 & 14873 \\
        \midrule
        \multirow{6}{*}{\rotatebox{90}{\footnotesize SP}} &
        Comparative Synonym   & 21,259 & 9,037 & 1,271 \\
        & Paraphrasing   & 92,761 & 38,374 & 5,323 \\
        & Noun Synonym   & 119,301 & 49,977 & 6,850 \\
        & Number Substitution   & 1,443 & 678 & 95 \\
        & Pronoun Substitution   & 44,435 & 19,025 & 2,620 \\
        & Verb Synonym   & 45,612 & 19,384 & 2,622 \\
        \midrule
        & Total SP & 324811 & 136475 & 18781 \\
        \bottomrule 
    \end{tabular}
    }
    \caption{Number of SISP-transformed samples generated per category for the VQA Yes-No dataset.}
    \label{tab:vqa_sisp_dataset_split}
\end{table}
In Tables~\ref{tab:nlvr2_sisp_dataset_split},~\ref{tab:violin_sisp_dataset_split},~\ref{tab:vqa_sisp_dataset_split}, we show the number of SISP-transformed samples generated for the test sets of NLVR$^2$, VIOLIN and VQA Yes/No respectively.
While we generate samples exhaustively for each category of transformation, during training these are sampled according to the proportion of augmented samples $T$, using three sampling strategies -- naive data augmentation, SW-SDRO or GW-SDRO.
On average, we obtain $5.69$ SI samples and $5.65$ SP samples per original sample for the NLVR$^2$ dataset, 
$11.14$ SI samples and $10.83$ SP samples for VIOLIN, and $2.75$ SI samples and $3.5$ SP sample for the VQA-Yes/No subset.

\subsection{Data Generation}
\begin{figure*}
    \centering
    \includegraphics[width=\linewidth]{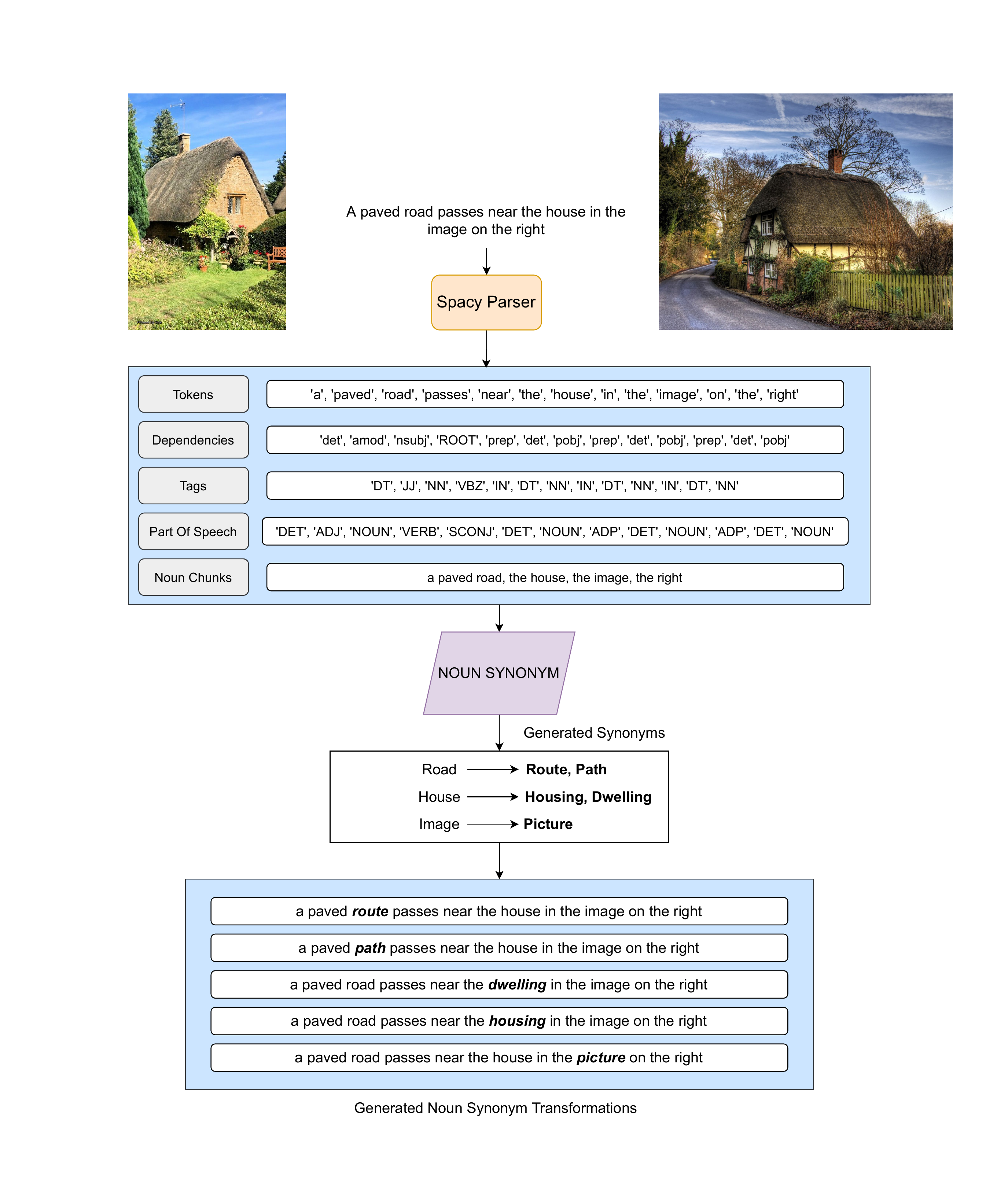}
    \caption{Illustration of the work-flow for generating SISP-transformed versions of input sentences. A Semantics-Preserving (SP) transformation is shown above.}
    \label{fig:sisp_sp}
\end{figure*}

\begin{figure*}
    \centering
    \includegraphics[width=\linewidth]{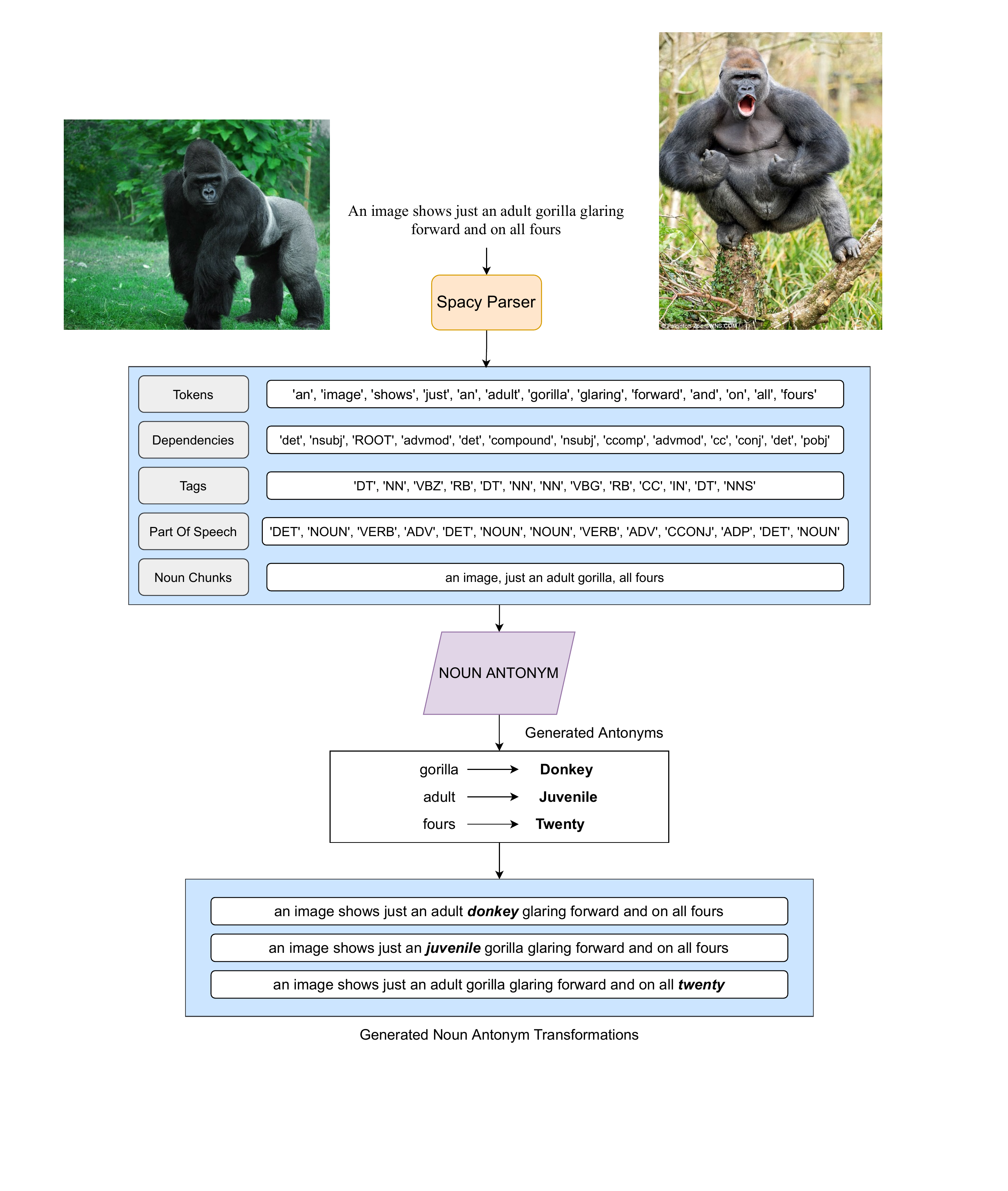}
    \caption{Illustration of the work-flow for generating SISP-transformed versions of input sentences. A Semantics-Inverting (SI) transformation is shown above.}
    \label{fig:sisp_si}
\end{figure*}

Figures~\ref{fig:sisp_sp} and~\ref{fig:sisp_si} show flowcharts for our SISP transformation process for Semantics Preserving (SP) and Semantics Inverting (SI) respectively.
For each image-sentence pair, the sentence is parsed using Spacy~\cite{spacy} into tokens, dependencies, POS-tags, and noun chunks.
Using this, each SISP function (for instance ``Noun Synonym'') generates insertions, deletions, substitutions, or paraphrasing as shown.

\begin{table}
    \centering
    \footnotesize
    % \resizebox{\linewidth}{!}{
    \begin{tabular}{@{}l ccccc@{}}
        \toprule
        \multirow{2}{*}{\textbf{Category}} & \multicolumn{5}{c}{\textbf{Fidelity Metrics}} \\
        \cmidrule{2-6}
         & LC & GC & VG & SC & Avg. \\
        \midrule
        SP  & 73.33 & 80.00 & 96.67 & 70.00 & 80.00\\
        SI  & 71.15 & 67.31 & 96.15 & 57.69 & 73.08\\
        All & 71.95 & 71.95 & 96.34 & 62.20 & 75.10\\
        \bottomrule
    \end{tabular}
    % }
    \caption{Human validation of our SISP transforms split according to the category of transformation.}
    \label{tab:human_sp_si}
\end{table}
\begin{table}
    \centering
    \footnotesize
    % \resizebox{\linewidth}{!}{
    \begin{tabular}{@{}l ccccc@{}}
        \toprule
        \multirow{2}{*}{\textbf{GT Label}} & \multicolumn{5}{c}{\textbf{Fidelity Metrics}} \\
        \cmidrule{2-6}
         & LC & GC & VG & SC & Avg. \\
        \midrule
        \texttt{True}  & 70.69 & 72.41 & 98.28 & 56.90 & 74.59 \\
        \texttt{False} & 75.00 & 70.83 & 91.67 & 75.00 & 78.13 \\
        All            & 71.95 & 71.95 & 96.34 & 62.20 & 75.10 \\
        \bottomrule
    \end{tabular}
    % }
    \caption{Human validation of our SISP transforms split according to the GT label of the original sample.}
    \label{tab:human_pos_neg}
\end{table}
\begin{table}[t]
    \centering
    \footnotesize
    % \resizebox{\linewidth}{!}{
    \begin{tabular}{@{}l ccccc@{}}
        \toprule
        \multirow{2}{*}{\textbf{GT Label}} & \multicolumn{5}{c}{\textbf{Fidelity Metrics}} \\
        \cmidrule{2-6}
         & LC & GC & VG & SC & Avg. \\
        \midrule
        SP(\texttt{True})   & 55.55 & 83.33 & 100.0 & 66.67 & 76.39\\
        SI(\texttt{True})   & 77.50 & 67.50 & 97.50 & 52.50 & 73.75\\
        SP(\texttt{False})  & 100.0 & 75.00 & 91.67 & 75.00 & 85.42\\
        SI(\texttt{False})  & 50.00 & 66.67 & 91.67 & 75.00 & 70.83\\
        \midrule
        All            & 71.95 & 71.95 & 96.34 & 62.20 & 75.10 \\
        \bottomrule
    \end{tabular}
    % }
    \caption{Human validation of our SISP transforms split according to the GT label of the original sample.}
    \label{tab:human_4way}
\end{table}

\subsection{Transformation Fidelity}
\begin{figure*}
    \centering
    \includegraphics[width=\linewidth]{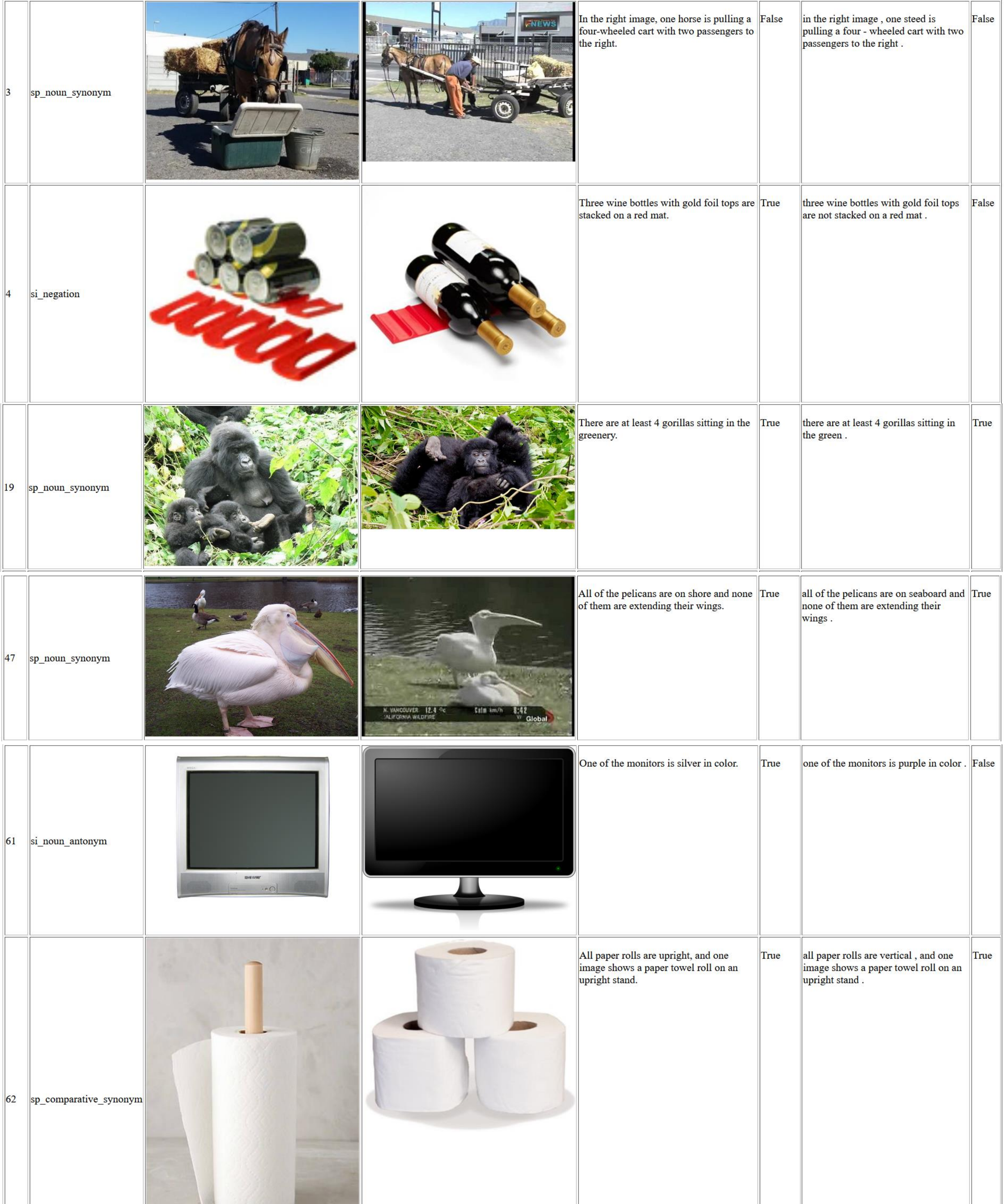}
    \caption{Snapshot of a SISP example being evaluated by human subjects.
    Columns from left to right: sample-ID, SISP-tag, Left Image, Right Image, Original Sentence, Original Label, New Sentence, New Label.}
    \label{fig:human_view}
\end{figure*}

For each of the 13 SISP categories, we sampled 100 SISP-transformed examples from NLVR$^2$, thus giving us a total of 1300 samples.
We employed 10 human subjects to evaluate the quality of SISP-transformed sentences.
These human subjects were all proficient in English and at the time of the study were enrolled in graduate programs in an English-speaking country.
The subjects were shown samples with the original images, sentences, and labels, as well as the new sentence and new label as shown in Figure~\ref{fig:human_view}.
These subjects evaluated each sample with a binary (0/1) score, according to $4$ metrics described below, along with an average ``Transformation Fidelity'':
\begin{enumerate}[noitemsep]
    \item Label Correctness (LC) -- \textit{Is the new label correct for the new sentence?}
    \item Grammatical Correctness (GC) --
    \textit{Does the sentence appear to be grammatically correct?}
    \item Visual Grounding (VG) -- \textit{Does the sentence refer to at-least one visual entity from the image?}
    \item Semantic Correctness (SC) -- \textit{Is the sentence semantically sound and  not absurd?}
\end{enumerate}

The subjects were asked to view each sample and rate the new sentence and label on a binary scale for each of the four metrics.
A snapshot of the interface used for the study as viewed by the human subjects is shown in Figure~\ref{fig:human_view}.
Results are shown in Table~\ref{tab:human_sp_si} -- split by the category of SISP transformation and in Table~\ref{tab:human_pos_neg} -- split by the ground-truth label of the original sample.
Overall, our SISP transformed test set for NLVR$^2$ was rated at an average fidelity of $75.10\%$. 
It can be observed that on average, SP samples were rated to have higher average fidelity than SI samples, and False samples higher than True samples.

We also split the ratings (2 SISP categories and 2 labels: $2\times2$) and show results in Table~\ref{tab:human_4way}.
Overall, $SP(False)$ has the highest average fidelity, and $SI(False)$ has the lowest.
LC (label correctness) for SI transformations of False statements is only $50\%$, probably because the inversion of a False statement using template-based methods may not always result in a True statement.
On the other hand an SP transformation of a False statement remains False and got $100\%$ LC.
It is surprising to observe that LC for SP transformations of True statements is low.
$SP(True)$ received the highest GC and VG ratings, but low SC and LC ratings.
VG ratings for all categories were consistently high.

\clearpage
\clearpage

\end{document}